\documentclass[10pt,journal,compsoc]{IEEEtran}
\usepackage[pagebackref,breaklinks,colorlinks]{hyperref}
\usepackage{booktabs}
\usepackage{graphicx}
\usepackage{amsmath}
\usepackage{amssymb}
\usepackage{booktabs}
\usepackage{multirow}
\usepackage{algorithm}
\usepackage{algorithmic}
\usepackage{multirow}
\usepackage{tabularx}
\usepackage{amsfonts,bm}
\usepackage{threeparttable}
\usepackage{xspace}
\usepackage{xcolor}
\usepackage{amsmath}
\usepackage{upgreek}
\usepackage{amssymb}
\usepackage{pifont}%
\newcommand{\cmark}{\ding{51}}%
\usepackage{multirow}
\usepackage{makecell}
\usepackage{color}
\usepackage{graphicx}
\usepackage{colortbl}
\usepackage{wrapfig}
\usepackage{soul}
\usepackage{ragged2e}

\usepackage[first=0,last=9]{lcg}
\definecolor{Gray}{gray}{0.86}

\newcommand{\xmark}{\ding{55}}%
\newcolumntype{x}[1]{>{\centering\arraybackslash}p{#1pt}}
\newcolumntype{y}[1]{>{\raggedright\arraybackslash}p{#1pt}}
\newcolumntype{z}[1]{>{\raggedleft\arraybackslash}p{#1pt}}

\ifCLASSOPTIONcompsoc
  \usepackage[caption=false,font=footnotesize,labelfont=sf,textfont=sf]{subfig}
\else
  \usepackage[caption=false,font=footnotesize]{subfig}
\fi

\makeatletter
\def\hlinew#1{%
  \noalign{\ifnum0=`}\fi\hrule \@height #1 \futurelet
   \reserved@a\@xhline}
\makeatother%

\title{Integrally Pre-Trained Transformer Pyramid Networks}
\definecolor{orange}{RGB}{255,127,0}

\begin{document}

% \markboth{IEEE TRANSACTIONS ON PATTERN ANALYSIS AND MACHINE INTELLIGENCE UNDER REVIEW}
% {Shell \MakeLowercase{\textit{et al.}}: Bare Demo of IEEEtran.cls for IEEE Journals}
% \markboth{IEEE TRANSACTIONS ON PATTERN ANALYSIS AND MACHINE INTELLIGENCE UNDER REVIEW}
% {Shell \MakeLowercase{\textit{et al.}}: Bare Demo of IEEEtran.cls for IEEE Journals}

\title{Fast-iTPN: Integrally Pre-Trained Transformer Pyramid Network with Token Migration}

\author{Yunjie Tian, Lingxi Xie, Jihao Qiu, Jianbin Jiao, Yaowei Wang, Qi Tian, Qixiang Ye
\IEEEcompsocitemizethanks{\IEEEcompsocthanksitem Y. Tian, J. Qiu, J. Jiao and Q. Ye are with the School of Electronic, Electrical and Communication Engineering, University of Chinese Academy of Sciences, Beijing, 101408, China. E-mail: \{tianyunjie19, qiujihao19\}@mails.ucas.ac.cn, \{jiaojb, qxye\}@ucas.ac.cn. Q. Ye is the corresponding author. \protect
\IEEEcompsocthanksitem L. Xie and Qi Tian are with Huawei Inc., China. Email: \{198808xc@gmail.com, zxphistory@gmail.com, tian.qi1@huawei.com\}. \protect
\IEEEcompsocthanksitem Y. Wang is with Peng Cheng Laboratory, Shenzhen, China, 100013. Email: \{wangyw@pcl.ac.cn\}.
}% <-this % stops an unwanted space
}
\IEEEtitleabstractindextext{%
\begin{abstract}
\justifying
We propose integrally pre-trained transformer pyramid network (iTPN), towards jointly optimizing the network backbone and the neck, so that transfer gap between representation models and downstream tasks is minimal. 
iTPN is born with two elaborated designs: 1) The first pre-trained feature pyramid upon vision transformer (ViT). 2) Multi-stage supervision to the feature pyramid using masked feature modeling (MFM) . 
\textcolor{black}{iTPN is updated to Fast-iTPN, reducing computational memory overhead and accelerating inference through two flexible designs. 1) Token migration: dropping redundant tokens of the backbone while replenishing them in the feature pyramid without attention operations. 2) Token gathering: reducing computation cost caused by global attention by introducing few gathering tokens.
\textcolor{black}The base/large-level Fast-iTPN achieve \textbf{88.75\%}/\textbf{89.5\%} top-1 accuracy on ImageNet-1K. With $1\times$ training schedule using DINO, the base/large-level Fast-iTPN achieves \textbf{58.4\%}/\textbf{58.8\%} box AP on COCO object detection, and a \textbf{57.5\%}/\textbf{58.7\%} mIoU on ADE20K semantic segmentation using MaskDINO.
\textcolor{black}Fast-iTPN can accelerate the inference procedure by up to \textbf{70\%}, with negligible performance loss, demonstrating the potential to be a powerful backbone for downstream vision tasks. }
The code is available at 
{\color{magenta}github.com/sunsmarterjie/iTPN}.
\end{abstract}
\begin{IEEEkeywords}
Integrally Pre-trained, Transformer Pyramid Networks, Masked Image Modeling, Downstream Generalization
\end{IEEEkeywords}
}

\maketitle
%%%%%%%%%%%%%%%%%%%%%%%%%%%%%%%%%%%%%%%%%%%%%%%%%%%%%%%%%%%%%%%%%%%%%%%%%%%%%%%
\section{Introduction}
\label{sec:intro}

\IEEEPARstart{I}{n} recent years, we have witnessed two major progresses in vision models, namely, the ViT architecture~\cite{ViT2021} as network backbone and the masked image modeling (MIM) method~\cite{bao2021beit, MAE2022,xie2022simmim} for model pre-training. Combining these two techniques yields a generalized pipeline that achieves state-of-the-art in a wide range of downstream tasks, including image classification, object detection, and instance/semantic segmentation.

Despite the progress, one remaining challenge of this pipeline is the transfer gap between upstream pre-training and downstream fine-tuning. From this point of view, we argue that downstream tasks, particularly fine-scaled recognition (\textit{e.g.}, detection and segmentation), require hierarchical features. However, most pre-training tasks (\textit{e.g.}, BEiT~\cite{bao2021beit} and MAE~\cite{MAE2022}) were built upon plain ViTs. Even if hierarchical ViTs have been used (\textit{e.g.}, in SimMIM~\cite{xie2022simmim}, ConvMAE~\cite{convmae}, and GreenMIM~\cite{huang2022green}), the pre-training task only affects the backbone but leaves the neck (\textit{e.g.}, a feature pyramid) un-trained. This brings risks to downstream tasks as finetuning starts with a randomly initialized neck which is not guaranteed to cooperate with the pre-trained backbone.

\begin{figure}[t]
\centering
\includegraphics[width=1\linewidth]{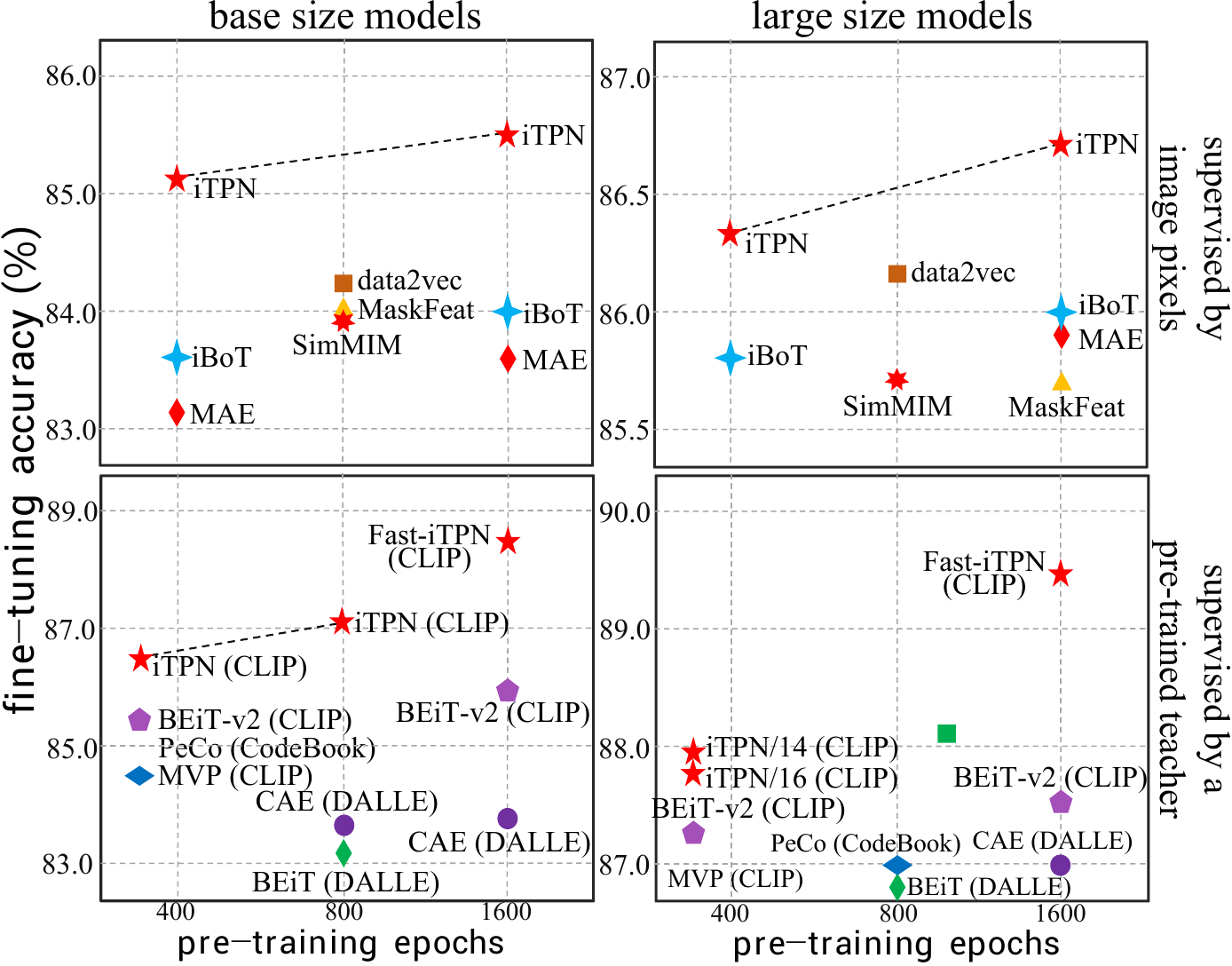}
\caption{Comparison of ImageNet-1K classification accuracy. iTPN/Fast-iTPN shows significant advantages over prior models, either using pixel supervision (upper) or knowledge distillation from a teacher (lower).}
\label{fig:overall}
\end{figure}

In this study, we propose integrally pre-trained pyramid transformer networks (iTPNs) to alleviate the risk. We establish the baseline with HiViT~\cite{hivit}, an MIM-friendly hierarchical transformer, and equip it with a feature pyramid. To jointly optimize the backbone (HiViT) and the neck (feature pyramid), we make two-fold technical contributions. \textbf{First}, we unify the upstream and downstream necks by inserting a feature pyramid into the pre-training stage (for reconstruction) and reusing the weights in the fine-tuning stage (for recognition). \textbf{Second}, to better pre-train the feature pyramid, we propose masked feature modeling (MFM) that 1) computes intermediate targets by feeding the original image into a moving-averaged backbone, and 2) uses the output of each pyramid stage to reconstruct the intermediate targets. MFM is complementary to MIM and improves the accuracy of both reconstruction and recognition. MFM can also be adapted to absorb knowledge from a pre-trained teacher (\textit{e.g.}, CLIP~\cite{clip}) towards higher performance, Figure~\ref{fig:overall}.

When using the hierarchical architecture and feature pyramid, the computational cost caused by global self-attention across tokens is aggregated. To alleviate this issue, we upgrade iTPN to Fast-iTPN by introducing two flexible designs: 1) Token migration: dropping redundant tokens from the backbone according to a similarity metric, and replenishing the dropped tokens to the feature pyramid, where there is no self-attention operations. 2) 
\textcolor{black}{Token gathering: introducing gathering tokens to aggregate global information from all windows so that global attention can be replaced with window attention, which significantly accelerates the inference with negligible performance cost.}
% Token gathering: introducing global information gathering tokens so that the global attention can be replaced with window attention.
With these designs, Fast-iTPNs achieves $70\%$ acceleration with minimal performance loss.

On ImageNet-1K, iTPN shows significant advantages, implying that the backbone itself becomes stronger when being jointly optimized with the neck. For example, the \textsf{base}/\textsf{large}-level iTPN reports an $\mathbf{86.2}\%$/$\mathbf{87.8}\%$ top-1 classification accuracy, beating the previous best record by $0.7\%$/$0.5\%$, which is not small as it seems in such a fierce competition. 
With upgraded training strategies, the Fast-iTPN-B/-L models boost the performance to \textbf{88.75\%/89.5\%} (without acceleration).
The iTPN series demonstrate advantages on downstream vision tasks. On COCO and ADE20K, iTPN largely benefits from the pre-trained feature pyramid. For example, the \textsf{base}/\textsf{large}-level iTPN reports a $\mathbf{53.2}\%$/$\mathbf{55.6}\%$ box AP on COCO ($1\times$ schedule, Mask R-CNN) and a $\mathbf{54.7}\%$/$\mathbf{57.7}\%$ mIoU on ADE20K (UPerNet). 
The Fast-iTPN-B model reports \textbf{61.8} mAP on COCO and \textbf{58.2} mIoU on ADE20K (pre-trained on Objects365~\cite{shao2019objects365}), surpassing the competitors by large margins. 
In diagnostic experiments, we show that iTPN enjoys 1) a lower reconstruction error in MIM pre-training; and 2) a faster convergence speed in downstream fine-tuning -- this validates that shrinking the transfer gap benefits both upstream and downstream parts.

The contributions of this study include:
\begin{itemize}
    \item We propose an integral pre-training feature pyramid network (iTPN), which, beyond setting new state-of-the-art, enlightens an important future research direction -- unifying upstream pre-training and downstream fine-tuning to shrink the transfer gap.
    \item We update iTPN to Fast-iTPN with upgraded training strategies. Fast-iTPN models report better performance across all the tested tasks. Furthermore, by introducing token migration and token gathering strategies, enabling Fast-iTPN to report faster inference speed and less memory cost, fascinating the downstream tasks friendly. 
    \item Under comparable computational cost, the iTPN/Fast-iTPN series outperforms existing vision models by significant margins in downstream tasks including image classification, object detection, and semantic/instance segmentation, demonstrating the potential to be a versatile foundation model.
\end{itemize}

\begin{figure*}[t]
\centering
\includegraphics[width=1\linewidth]{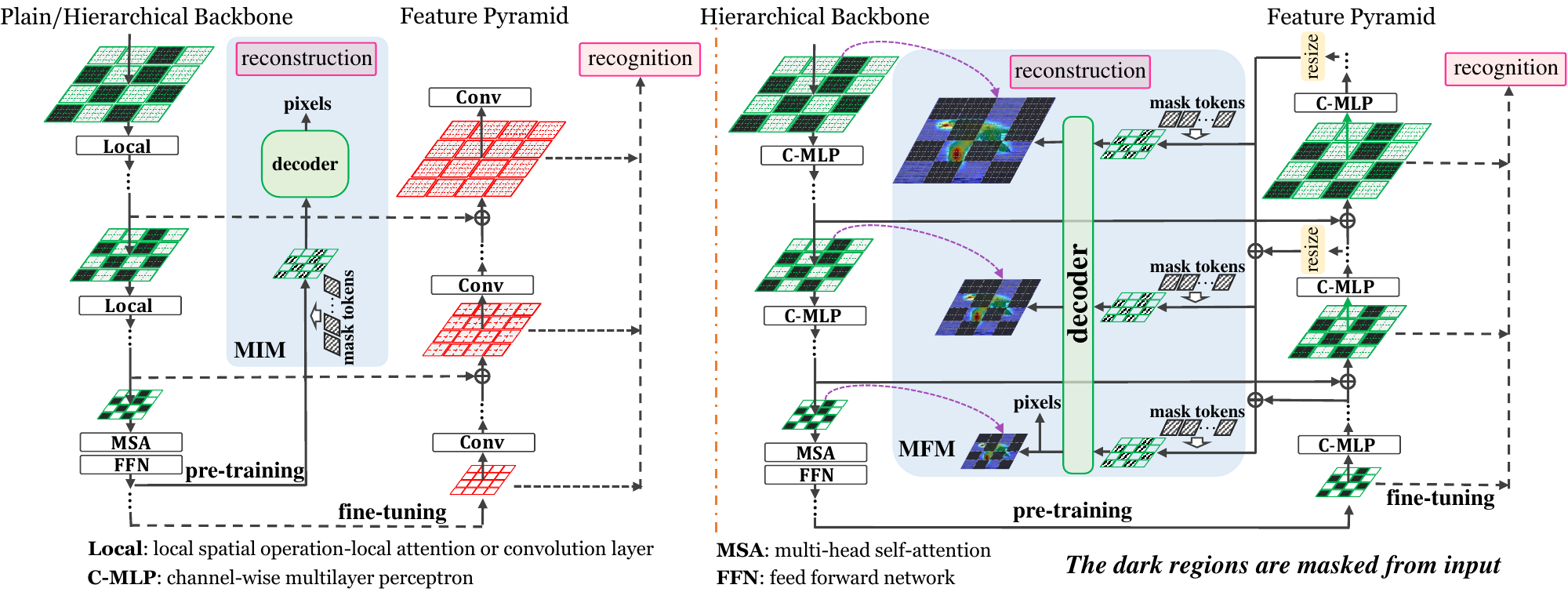}
\caption{The comparison between a conventional pre-training (left) and the proposed integral pre-training framework (right). We use a feature pyramid as the unified \textbf{neck} module and apply masked feature modeling to pre-train the feature pyramid. The blocks in green and red respectively indicate that the network weights are pre-trained and un-trained (\textit{i.e.}, randomly initialized for fine-tuning).}
\label{fig:framework}
\end{figure*}

\section{Related Work}
\label{sec:related}

In the deep learning era~\cite{lecun2015deep}, vision models are mostly built upon deep neural networks, $e.g.$, CNNs~\cite{Resnet2016,AlexNet2012,EfficientNet2019} and ViTs~\cite{ViT2021,Swin2021,hivit,pvt}. This study focuses on the ViTs that were transplanted from the natural language processing field~\cite{Attention2017}. The core idea is to extract visual features by treating each image patch as a token and computing self-attentions among them.

\subsection{Hierarchical Vision Transformers}

The \textbf{vanilla ViTs} appeared in a plain form~\cite{ViT2021,yuan2021tokens,zhou2021deepvit, chen2022adaptformer, li2022efficientformer} where, throughout the backbone, the number of tokens keeps a constant and the attention among these tokens is symmetric. To compensate for the inductive priors in computer vision, the community designed \textbf{hierarchical ViTs}~\cite{pvt,wu2021cvt,dai2021coatnet, Swin2021, hivit, tian2023integrally} which allow the number of tokens to gradually decrease throughout the backbone, \textit{i.e.}, similar as in CNNs. Other design principles were also inherited, such as introducing convolution to ViTs so that the relationship between neighborhood tokens is better formulated~\cite{mvit, pvt, dai2021coatnet, container, mvit_v2,wang2022pvt}, interacting between hybrid information~\cite{dai2021coatnet}, using window~\cite{Swin2021,dong2022cswin,hivit} or local~\cite{yuan2021volo} self-attentions to replace global self-attentions, adjusting the geometry for local-global interaction~\cite{yang2021focal}, decomposing self-attentions~\cite{MaxViT2022}, and so on. It was shown that hierarchical ViTs offer high-quality, multi-level visual features that easily cooperate with a neck module (for feature aggregation, \textit{e.g.}, a feature pyramid~\cite{FPN2017}) and benefit downstream tasks.

Most hierarchical ViTs~\cite{wang2022pvt, wu2021cvt, pvt, dai2021coatnet} borrows design principles from the feature pyramid network (FPN)~\cite{FPN2017}, self-attention geometry~\cite{yang2021focal}, and hybrid network architectures~\cite{Conformer2021}. One effective way is integrating hierarchical convolution layers with transformers~\cite{container, dai2021coatnet, Conformer2021}. The other leverages attention at multiple scales to enhance global attention~\cite{dong2022cswin, yuan2021volo}. To operate high-resolution transformer features efficiently, global attention operations are typically replaced by local/window attention operations~\cite{dong2022cswin} or decomposed to axis attentions~\cite{MaxViT2022}. 

\subsection{Vision Transformer Pre-Training}

The continuous growth of vision data calls for visual pre-training, in particular, self-supervised learning that learns generic representations from unlabeled data. At the core of self-supervised learning lies a pretext task, \textit{i.e.}, an unsupervised learning objective that the model pursues. The community started with preliminary pretext tasks such as geometry-based tasks (\textit{e.g.}, determining the relative position between image patches~\cite{doersch2015unsupervised, jigsaw} or the transformation applied to an image~\cite{rotation}), and generation-based tasks (\textit{e.g.}, recovering the removed contents~\cite{inpainting} or attributes~\cite{colorization, tian2022beyond, tian2021semantic} of an image), but these methods suffer unsatisfying accuracy (\textit{i.e.}, trailed by fully-supervised pre-training significantly) when transferred to downstream tasks. The situation was changed when new pretext tasks were introduced, in particular, contrastive learning~\cite{grill2020bootstrap,he2020momentum,chen2020simple,grill2020bootstrap,caron2021emerging,swav,pixpro} and \textbf{masked image modeling} (MIM)~\cite{bao2021beit,MAE2022,xie2022simmim, li2022semmae, li2022mc, assran2022masked, chen2022sdae}, where the latter is yet another type of generation-based learning objective.

This study focuses on MIM, which takes advantage of ViTs that formulate each image patch as a token. The tokens can be arbitrarily masked (discarded from the input data) and the learning objective is to recover the masked contents at the pixel level~\cite{MAE2022, xie2022simmim, tian2022beyond, tian2023integrally}, the feature level~\cite{bao2021beit, maskfeat, zhang2022cae}, or in the frequency space~\cite{devil}. MIM has shown an important property named scalability, \textit{i.e.}, augmenting the amount of pre-training data (\textit{e.g.}, from ImageNet-1K to ImageNet-22K) and/or increasing the model size (\textit{e.g.}, from the base level to the large or huge level) can boost the downstream performance~\cite{MAE2022, chen2022context}, which aligns with the observations in language modeling~\cite{devlin2018bert, gpt3}. Recent large vision models~\cite{fang2023eva, fang2023eva02, wang2023internimage} also adopt MIM for pre-training, which significantly boosts the performance.

\subsection{Efficient Vision Transformers}

When applying ViTs to downstream tasks, decreasing the computational cost brought by global self-attention is crucial. PVT~\cite{pvt} introduced multi-resolution stages and down-sampled features to reduce computational costs. Swin Transformer~\cite{Swin2021} computed self-attention operation within local windows. To address the trade-off between lacking communication in non-overlapping windows and introducing additional memory/computation costs in overlapping windows, Swin Transformer modified the window partition style to implicitly establish connections between non-overlapping windows. Axial self-attention ~\cite{ho2019axial} and criss-cross attention~\cite{huang2019ccnet} suggested the calculation of attention within stripe windows along the horizontal and/or vertical axes. CSWin Transformer~\cite{dong2022cswin} developed a Cross-Shaped Window self-attention mechanism to compute self-attention in the horizontal and vertical stripes in parallel that form a cross-shaped window. Subsequent works~\cite{chu2021twins,huang2021shuffle,yu2021glance} established local-global relations by connecting local regions to avoid the time-consuming global self-attention.

The most related work to this study is HiViT~\cite{hivit} which builds hierarchy features using two MLP-based stages before the global attention stage so that convolution operations or window attention is not required. This guarantees the computational efficiency and compatibility with MIM. HiViT is selected as iTPN's backbone in this study. To further improve the efficiency, we propose to drop tokens of the HiViT backbone, which is decelerated by global attention operations, while replenishing these dropped tokens to the feature pyramid, which is computationally efficient as it does not involve global attention.

\section{iTPN}
\label{sec:method}

\subsection{Motivation: Integral Pre-Training}
\label{sec:method:motivation}

Denote $\mathcal{D}^\mathrm{pt}=\{\mathbf{x}_n^\mathrm{pt}\}_{n=1}^N$ an image dataset, where $N$ is the image number. The samples have no category labels. The fine-tuning phase involves another dataset $\mathcal{D}^\mathrm{ft}=\{\mathbf{x}_m^\mathrm{ft},\mathbf{y}_m^\mathrm{ft}\}_{m=1}^M$, where $M$ is the number of samples and $\mathbf{y}_m^\mathrm{ft}$ is the semantic label of $\mathbf{x}_m^\mathrm{ft}$.
The target model is composed of a backbone, a neck, and a head\footnote{Following the conventional definition, the neck is used for multi-stage feature aggregation (\textit{e.g.}, a feature pyramid~\cite{FPN2017}) while the head (\textit{e.g.}, a linear classifier) is used for prediction,} denoted as $f(\cdot;\boldsymbol{\theta})$, $g(\cdot;\boldsymbol{\phi})$, $h(\cdot;\boldsymbol{\psi})$, respectively. $\boldsymbol{\theta}$, $\boldsymbol{\phi}$, $\boldsymbol{\psi}$ are learnable parameters omitted for simplicity. $f(\cdot)$ takes $\mathbf{x}$ as input, while $g(\cdot)$ and $h(\cdot)$ works on the outputs of $f(\cdot)$ and $g(\cdot)$, \textit{i.e.}, the entire function is $h(g(f(\mathbf{x};\boldsymbol{\theta});\boldsymbol{\phi});\boldsymbol{\psi})$.

Throughout this paper, the pre-training task is masked image modeling (MIM) and the fine-tuning tasks include image classification, object detection, and instance/semantic segmentation. These tasks share the same backbone, but use different necks and heads. The self-supervised pre-training and fine-tuning objective functions are formulated as
\begin{equation}
\begin{aligned}
&\min\mathbb{E}_{\mathcal{D}^\mathrm{pt}}\|\mathbf{x}_n^\mathrm{pt}-h^\mathrm{pt}(g^\mathrm{pt}(f(\mathbf{x}_n^\mathrm{pt};\boldsymbol{\theta}),\boldsymbol{\phi}^\mathrm{pt}),\boldsymbol{\psi}^\mathrm{pt})\|,\\
&\min\mathbb{E}_{\mathcal{D}^\mathrm{ft}}\|\mathbf{y}_m^\mathrm{ft}-h^\mathrm{ft}(g^\mathrm{ft}(f(\mathbf{x}_m^\mathrm{ft};\boldsymbol{\theta}),\boldsymbol{\phi}^\mathrm{ft}),\boldsymbol{\psi}^\mathrm{ft})\|.
\end{aligned}
\end{equation}
%{\color{red}where parameters are not shared between $\boldsymbol{\phi}^\mathrm{pt}$, $\boldsymbol{\phi}^\mathrm{ft}$ and $\boldsymbol{\psi}^\mathrm{pt}$, $\boldsymbol{\psi}^\mathrm{ft}$.} 

We argue that such a pipeline leads to a significant transfer gap between the pre-trained models and downstream tasks, which thereby brings two-fold drawbacks. \textbf{First}, the backbone parameters, $\boldsymbol{\theta}$, are not optimized towards being used for multi-level feature extraction. \textbf{Second}, the fine-tuning phase starts with optimizing a randomly initialized $\boldsymbol{\phi}^\mathrm{ft}$ and $\boldsymbol{\psi}^\mathrm{ft}$, which may slow down the training procedure and lead to unsatisfying recognition results. To alleviate the gap, we advocate for an integral framework that unifies $g^\mathrm{pt}(\cdot)$ and $g^\mathrm{ft}(\cdot)$, so that the pre-trained $\boldsymbol{\phi}^\mathrm{pt}$ is easily reused to be an initialization of $\boldsymbol{\phi}^\mathrm{ft}$, and thus only $\boldsymbol{\psi}^\mathrm{ft}$ is randomly initialized, Figure~\ref{fig:framework}.

\subsection{Unifying Reconstruction and Recognition}
\label{sec:method:unifying}

Suppose the backbone model consists of $S$ stages and each stage has multiple transformer blocks. Most often, the backbone (\textit{a.k.a} encoder) gradually down-samples the input signal and produces $S+1$ feature maps, as
\begin{equation}
f(\mathbf{x};\boldsymbol{\theta})=\mathbf{U}^0,\mathbf{U}^1,\ldots,\mathbf{U}^S,
\end{equation}
where $\mathbf{U}^0$ denotes the direct embedding of input, and a smaller superscript index indicates a stage closer to the input layer. Each feature map is composed of a set of tokens, $\mathbf{U}^s=\{\mathbf{u}_1^s,\mathbf{u}_2^s,\ldots,\mathbf{u}_{K^s}^s\}$, where $K^s$ is the number of tokens in the $s$-th feature map.

We show that $g^\mathrm{pt}(\cdot)$ and $g^\mathrm{ft}(\cdot)$ can share the same architecture and parameters because both of them start with $\mathbf{U}^S$ and gradually aggregate it with lower-level features. Accordingly, we write the neck part, as
\begin{equation}
\begin{aligned}
\mathbf{V}^S&=\mathbf{U}^S,\\
\mathbf{V}^s&=\mathbf{U}^s+g^s(\mathbf{V}^{s+1};\boldsymbol{\phi}^s),\quad 1\leqslant s<S,
\end{aligned}
\end{equation}
where $g^s(\cdot)$ up-samples $\mathbf{V}^{s+1}$ to fit the resolution of $\mathbf{V}^s$. Note that the learnable parameters ($\boldsymbol{\phi}$) are composed of a layer-wise parameter set, $\{\boldsymbol{\phi}^s\}$. When these parameters are reused in fine-tuning, we largely shrink the transfer gap: the only modules that remain individual between pre-training and fine-tuning are the heads (\textit{e.g.}, the decoder for MIM \textit{vs.} the Mask R-CNN head for detection).

Before entering the next part that discusses the loss terms, we remind the readers that other differences exist between pre-training and fine-tuning, while they do not impact the overall design of network architectures.
\begin{itemize}
\item MIM samples a random mask $\mathcal{M}$ and applies it to $\mathbf{x}$, \textit{i.e.}, $\mathbf{x}$ is replaced by $\mathbf{x}\odot\overline{\mathcal{M}}$. Consequently, all the backbone outputs, $\mathbf{U}^0,\ldots,\mathbf{U}^S$, do not contain the tokens with indices in $\mathcal{M}$, and so are $\mathbf{V}^1,\ldots,\mathbf{V}^S$. At the start of the decoder, $\mathbf{V}=\sum_{s=1}^S\mathbf{V}^s$ is complemented by adding dummy tokens to the masked indices and then fed into a decoder for image reconstruction.
\item The downstream fine-tuning procedure makes use of specific outputs of the decoder for different tasks. For image classification, $\mathbf{V}^S$ is used. For detection and segmentation, all of $\mathbf{V}^1,\ldots,\mathbf{V}^S$ are used.
\end{itemize}

\newcommand{\pl}{\!+\!}
\begin{table}[]
\centering
\setlength{\tabcolsep}{0.35cm}
\caption{A comparison between ViT, Swin, HiViT, and the proposed iTPN in terms of network configuration. We use $224\times224$ input size to calculate the FLOPs. $\dagger$: We add $4$ Stage-3 blocks to HiViT-B to keep the FLOPs of iTPN-B comparable to ViT.}
\setlength{\tabcolsep}{0.08cm}
\begin{tabular}{l|c|c|c|c}
\toprule
Model & ViT &Swin~\cite{Swin2021} & HiViT~\cite{hivit} & iTPN \\
\midrule
\multicolumn{4}{l}{\textit{\textsf{base}-level models}} \\
\# stages & $1$  &$4$  & $3$ & $3$ \\
\# blocks & $12$ &$2\pl2\pl18\pl2$ & $3\pl3\pl20$ & $3\pl3\pl24^\dagger$ \\
% \# dec. blocks & $8$ & $6$ & $4\pl4\pl4$ \\
Params (M) & $86$  & $88$ & $66$ & $79$ \\
FLOPs (G) & $17.5$  & $15.4$ & $15.9$ & $17.8$ \\
% Time/ep (h) & $?.?$ & $?.?$ & $?.?$ & $?.?$ \\
\midrule
\multicolumn{4}{l}{\textit{\textsf{large}-level models}} \\
\# stages & $1$ &$4 $ & $3$ & $3$ \\
\# blocks & $24$  &$2\pl2\pl18\pl2$ & $2\pl2\pl40$ & $2\pl2\pl40$ \\
% \# dec. blocks & $8$ & $6$ & $4\pl4\pl4$ \\
Params (M) & $307$ &$197$ & $288$ & $288$ \\
FLOPs (G) & $61.3$ & $34.5$ & $61.2$ & $61.2$ \\
% Time/ep (h) & $?.?$ & $?.?$ & $?.?$ & $?.?$ \\
\bottomrule
\end{tabular}
\label{tab:config}
\end{table}

\subsection{Masked Feature Modeling}
\label{sec:method:mfm}

We first inherit the reconstruction loss from MIM that minimizes $\|\mathbf{x}-h^{\mathrm{pt},0}(\mathbf{V};\boldsymbol{\psi}^{\mathrm{pt},0})\|$, where $h^{\mathrm{pt},0}(\cdot)$ involves a few transformer blocks that reconstruct the original image from $\mathbf{V}=\sum_{s=1}^S\mathbf{V}^s$. To capture multi-stage features, we add a reconstruction head to each stage, termed $h^{\mathrm{pt},s}(\cdot;\boldsymbol{\psi}^{\mathrm{pt},s})$, and optimize the multi-stage loss, defined as
\begin{equation}
\mathcal{L}=\underbrace{\|\mathbf{x}-h^{\mathrm{pt},0}(\mathbf{V})\|}_\mathrm{image\ reconstruction}+\lambda\cdot\underbrace{\sum_{s=1}^S\|\mathbf{x}^s-h^{\mathrm{pt},s}(\mathbf{V}^s)\|}_\mathrm{feature\ reconstruction},
\end{equation}
where $\mathbf{x}^s$ is the expected output at the $s$-th decoder stage, and $\lambda=0.3$ is determined in a held-out validation set. Since the goal is to recover the masked features, we name the second term masked feature modeling (MFM) loss which complements the first term, the masked image modeling (MIM) loss. We illustrate MFM in Figure~\ref{fig:framework}.

\begin{figure*}[t]
\centering
\includegraphics[width=0.95\linewidth]{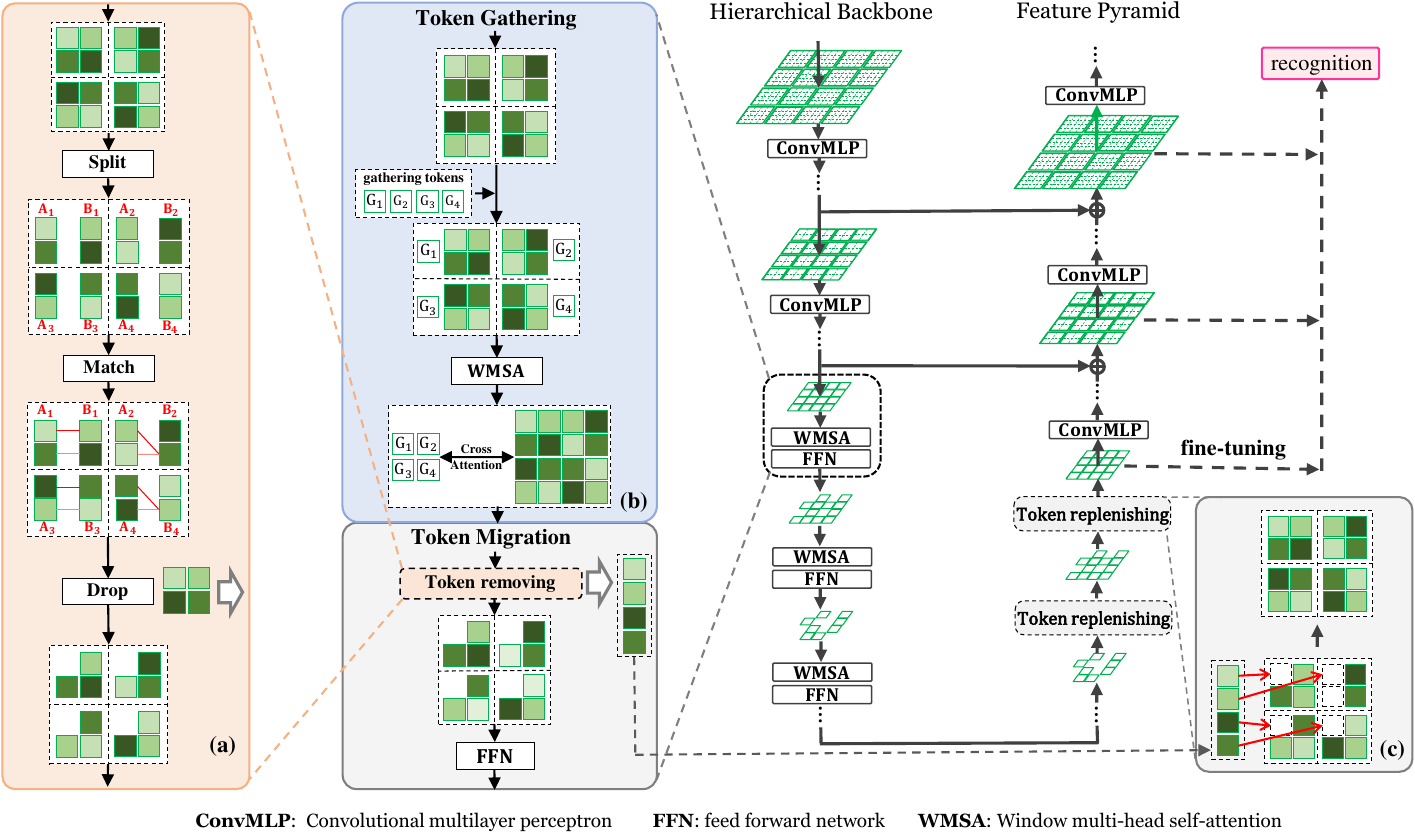}
\caption{\color{black}Fast-iTPN architecture during fine-tuning. Illustration of (a) token dropping from the backbone, (b) token gathering on the backbone, and (c) token replenishing to the feature pyramid.}
\label{fig:fast_itpn}
\end{figure*}

The remaining issue is to determine the intermediate reconstruction target, \textit{i.e.}, $\mathbf{x}^1,\ldots,\mathbf{x}^S$. We borrow the idea from knowledge distillation~\cite{hinton2015distilling} that makes use of a teacher backbone $\hat{f}^\mathrm{back}(\cdot)$ to generate the intermediate targets, \textit{i.e.}, $\hat{f}(\mathbf{x};\hat{\boldsymbol{\theta}})=\mathbf{x}^1,\ldots,\mathbf{x}^S$. The teacher model is chosen to be the moving-averaged~\cite{meanteacher} encoder (in this case, no external knowledge is introduced) or another pre-trained model (\textit{e.g.}, CLIP~\cite{clip}, as used by~\cite{mvp,fd_clip}, that was pre-trained on a large dataset of image-text pairs). In the former case, we only feed the masked patches ($\mathbf{x}\odot\mathcal{M}$, not the entire image) to the teacher model for acceleration. In the latter case, we follow BEiT~\cite{bao2021beit} to feed the entire image to the pre-trained CLIP model.

As a side comment, the benefits brought by integral pre-training and MFM are individual and complementary -- we shall reveal this point in the ablation (Table~\ref{tab:ablation2}).

\subsection{Implementation Detail}
\label{sec:method:details}

We build the system upon HiViT~\cite{hivit}, our proposed hierarchical vision transformer. HiViT simplified the Swin transformers~\cite{Swin2021} by (i) replacing early-stage shifted-window attentions with channel-wise multi-layer perceptrons (C-MLPs) and (ii) removing the $7\times7$ stage so that global attentions are computed on the $14\times14$ stage. With these improvements, when applied to MIM, HiViT allows the masked tokens to be directly discarded from input (by contrast, with Swin as the backbone, SimMIM~\cite{xie2022simmim} required the entire image to be used as input), saving $30\%$--$50\%$ computational costs and leading to better performance. 

\textcolor{black}{Without shifted-window attentions or convolution operations, we can leverage simple token merging in the backbone and token splitting operations in the neck to construct the first genuine Transformer Pyramid Network, which guarantees that overall iTPN architecture is friendly to masked image modeling.}

Table~\ref{tab:config} summarizes the configuration of iTPN. The computational complexity is comparable to that of ViT. We follow the convention to use $224\times224$ images during the pre-training. HiViT produces three stages ($S=3$) with spatial resolutions of $56\times56$, $28\times28$, and $14\times14$, respectively. An $S$-stage feature pyramid is built upon the backbone. We replace all convolutions in the feature pyramid with C-MLPs to avoid leaking information from visible patches to invisible patches. As we shall see in ablation (Section~\ref{exp:sec:analysis}), using C-MLP in the feature pyramid leads to consistent accuracy gain in various visual recognition tasks, and the improvement is complementary to that brought by MFM.

\textcolor{black}{Regarding MFM, we investigate two choices of the teacher model. (i) The first option involves computing the exponential moving average (EMA) of the online target model with a coefficient of $0.996$. We extract the supervision from the last layer of each stage so that for any $s$, $\mathbf{x}^s$ has the same spatial resolution as $\mathbf{V}^s$, and thus $h^s(\cdot)$ is a linear projection working on each token individually. (ii) The second option directly inherits a CLIP pre-trained model. Note that CLIP offers standard ViTs that do not produce multi-resolution feature maps. In this scenario, we unify the $S$ MFM terms into one by down-sampling all the feature maps to the lowest spatial resolution ($14\times14$), adding them together, and comparing the sum to the last-layer output of the CLIP model.}

\section{Fast-iTPN}
{\textcolor{black}{Majority of iTPN's computational cost overhead comes from the backbone, where global self-attention operations are performed across all tokens. To reduce the computational cost overhead, we propose two strategies: 1) Token migration: dropping redundant tokens on the backbone while replenishing these tokens to the feature pyramid. As the feature pyramid does not use any attention operation, token migration significantly reduces the computational cost and memory. 2) Token gathering: introducing gathering tokens to gather global information about tokens so that global attention can be replaced with window attention.}}
\textcolor{black}{These two acceleration strategies are only applied to the fine-tuning stage for downstream tasks. }

\subsection{Token Migration}
{\textcolor{black}{\textbf{Token Dropping.} As is known, images exhibit a notable characteristic, namely, information sparsity. In other words, in an image only a limited number of tokens/regions carry representative information, while the majority of regions are deemed redundant. This observation motivates our proposition: dropping similar tokens in the backbone to diminish computational cost while recalling the dropped tokens on the feature pyramid to guarantee the feature completeness.}}

Specifically, token dropping starts from the third stage of the backbone transformer, where self-attention mechanisms are employed. Mathematically,
given an image $\mathbf{x}_m^\mathrm{ft}$, $f(\mathbf{x}_m^\mathrm{ft};\boldsymbol{\theta})$ first outputs the first two feature maps.
\begin{equation}
f(\mathbf{x}_m^\mathrm{ft};\boldsymbol{\theta})^\mathrm{stage_{1, 2}}=\mathbf{U}^0,\mathbf{U}^1
\end{equation}
where the $\mathbf{U}^1=\{\mathbf{u}_1^1,\mathbf{u}_2^1,\ldots,\mathbf{u}_{K^1}^1\}$. The tokens in $\mathbf{u}_{K^1}^1$ are first migrated as noted below.
In each layer, Figure~\ref{fig:fast_itpn} (a), tokens within a window are divided into two distinct sets, $\mathcal{A}$ and $\mathcal{B}$. The pairwise similarity between tokens from sets $\mathcal{A}$ and $\mathcal{B}$ is computed as the mean of the attention map's K-matrix. The top-\textit{r} tokens of the largest similarity within set $\mathcal{A}$ are dropped from the backbone, and the token numbers and positions are stored in a sequence. \textit{r} is an experimentally determined hyper-parameter.
\textcolor{black}{
Denote the migration function as $\mathcal{M(\cdot)}$, the input for the first layer of the third stage can be represented as $M(\mathbf{u}_{K^1}^1)$, which outputs the feature maps of the remained tokens, feature maps and positions of the migrated tokens. The migrated feature maps and positions will be cast into a store sequence $\mathrm{ss(\cdot)}$. For $\mathit{l}$-th layer of the third stage, there is an input $M(\mathbf{u}_{l}^2)$ from the former layer, 
\begin{equation}
f(M(\mathbf{u}_{l}^2);\boldsymbol{\theta})_{l}^\mathrm{stage_{3}}=\mathbf{u}_{l+1}^2, \mathbf{u}_{l+1}^{\prime2}, \mathbf{pos}_{l+1}
\end{equation}
}
\textcolor{black}{
where $\mathbf{u}_{l+1}^{\prime2}$ is the migrated tokens and $\mathbf{pos}_{l+1}$ is the corresponding positions.}

{\textcolor{black}{
Intuitively, more migrated tokens lead to a larger performance drop and greater speed up. Experimental results show that substantial performance degradation is primarily observed when a substantial number of tokens are migrated, Figure~\ref{fig:perf_speed_trade}. The experimental results are consistent with that of ToMe~\cite{bolya2022token}. The difference lies in that we directly discard tokens while ToMe uses a token merging strategy. 
}}

{\textcolor{black}{\textbf{Token Replenishing.} 
The migrated backbone tokens in $\mathrm{ss(\cdot)}$ are replenished to the feature pyramid $g^\mathrm{ft}$ to guarantee feature completeness while mitigating the potential for significant performance degradation, Figure~\ref{fig:fast_itpn} (a).
Specifically, we retrieve the numbers and positions of previously migrated tokens from $\mathrm{ss(\cdot)}$.  These tokens are then replenished to the feature pyramid according to their numbers and positions on the backbone, Figure~\ref{fig:fast_itpn} (c).
}}

\begin{figure}[t]
\centering
\includegraphics[width=0.8\linewidth]{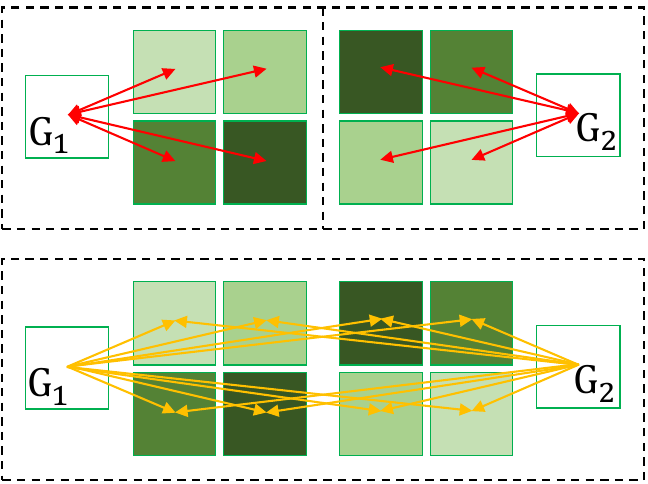}
\caption{\textcolor{black}{Token gathering. Top: global attention within the respective window for each gathering token. Bottom: integrating information from all gathering tokens to all tokens. }}
\label{fig:token_gathering}
\end{figure}

\begin{figure}[t]
\centering
\includegraphics[width=1\linewidth]{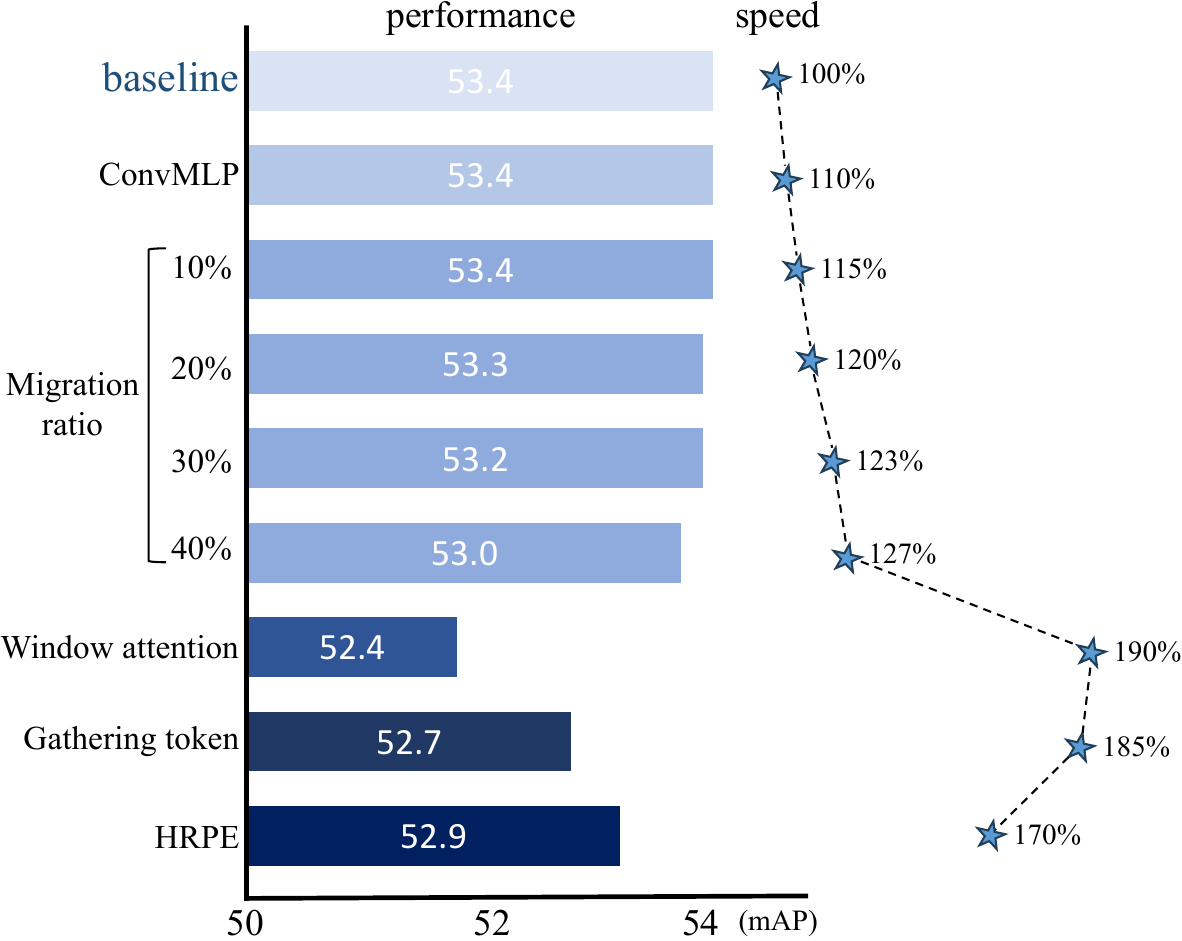}
\caption{\textcolor{black}{Speed-up of Fast-iTPN over iTPN. The experiments is done using Mask-RCNN for object detection on COCO by fine-tuning the pre-trained Fast-iTPN-B model for 12 epochs. Fast-iTPN achieves a speedup of approximately 20\% with minimal performance loss ($\sim$ 0.2) and a speedup of 70\% with a marginal decrease in mAP of 0.5\%. }}
\label{fig:perf_speed_trade}
\end{figure}

\begin{table}[]
\centering
\setlength{\tabcolsep}{0.5cm}
\caption{\textcolor{black}{Architecture modification of Fast-iTPN upon iTPN during pre-training.}}
\setlength{\tabcolsep}{0.08cm}
\begin{tabular}{l|c|c|c|c|c}
\toprule
\textcolor{black}{Model}   & \textcolor{black}{RPE}   &\textcolor{black}{subln} & \textcolor{black}{swiglu} & \textcolor{black}{MLP}  & \textcolor{black}{mlp ratio} \\
\midrule
\textcolor{black}{iTPN}      & \textcolor{black}{\cmark}    & \textcolor{black}{\xmark}   & \textcolor{black}{\xmark}   & \textcolor{black}{MLP}    & \textcolor{black}{4} \\
\textcolor{black}{Fast-iTPN} & \textcolor{black}{\xmark}    & \textcolor{black}{\cmark}   & \textcolor{black}{\cmark}   & \textcolor{black}{ConvMLP} & \textcolor{black}{3} \\
\bottomrule
\end{tabular}
\label{tab:config_fast_itpn}
\end{table}

\subsection{Token Gathering}

\textbf{Window Attention.}
This technique was applied in the Swin Transformer~\cite{Swin2021}, by partitioning image tokens to $K$ groups and computing attention within group tokens, termed window attention. Supposing the token count is $N$, the usage of window attention reduces the computational complexity from $O(N^2)$ to $O((\frac{N}{K})^2)$. Nevertheless, window attention reduces the receptive field, which often causes performance deterioration, particularly when handling large-size images.

\textbf{Global Information Gathering.}
To strike a favorable balance between performance and efficiency, we introduce a ``gathering token" for each window to gather global information. 
This kind of token orchestrates window attention within its designated window. As detailed in Figure~\ref{fig:token_gathering}, Fast-iTPN conducts global attention within each window together with an attached gathering token. 
These gathering tokens oversee the window-specific attention and then engage in cross-attention operations with the overarching token set to collect global information. 
As shown, with gathering tokens, the cross-attention operation exclusively focuses on computing the $q$ and $k$ values between the gathering tokens and all other tokens. 
This excludes the calculation of $q$ and $k$ values among all tokens themselves. 
%The global information within these window-leader tokens is transmitted into their respective windows.

%%%%%%%%%%%%%%%%%%%%%%%%%%%%%%%%%%%%%%%%%%%%%%%%%%%%%%%%%%%%%%%%%%%%
\subsection{Implementation Details}

Fast-iTPN inherits the pre-training strategies of iTPN but with some small modifications as shown in Table~\ref{tab:config_fast_itpn}. 
Specifically, we omitted relative positional embedding (RPE) during the pre-training phase, eliminating the need for its removal in tasks that do not necessitate its usage. 
Subln~\cite{ba2016layer} and swiglu~\cite{ramachandran2017searching} are also employed to enhance the performance of pre-trained models~\cite{fang2023eva02}.
Moreover, we have substituted the MLP layers employed in iTPN with $1\times1$ convolutional layers (ConvMLP) to expedite computation. The MLP ratio has been adjusted from 4 to 3 to balance parameters and FLOPs.

The efficiency advantage of Fast-iTPN compared to iTPN lies in the fine-tuning of dense downstream tasks. 
For these tasks, the input image sizes are often large, thereby, the usage of additional acceleration strategies is useful. 
During the fine-tuning phase, the pre-trained models will be directly transferred to the downstream tasks with two acceleration strategies that can be selectively applied. In the experiment, we will demonstrate the performance results without acceleration strategies and with different combinations of acceleration strategies.

\section{Experiment}
\label{sec:experiment}

\subsection{Settings}
iTPN is pre-trained on the ImageNet-1k dataset, a subset of ImageNet that contains 1.28M training images of 1,000 classes. The class labels are not used during the pre-training stage. Each training image is pre-processed into 224 × 224 and partitioned into 14 × 14 patches sized 16 × 16 pixels. Following MAE, a random subset of 75\% patches are masked from input, and the normalized pixels are preserved for reconstruction.

We use an AdamW optimizer with an initial learning rate of $\mathbf{1.5 \times 10}^{-4}$, a weight decay of 0.05, and a batch size of 4,096. The learning rate follows a cosine annealing schedule and the number of warm-up epochs is set to be 40. As mentioned above, MFM supervision may come from the moving-averaged online model or a pre-trained CLIP model. The numbers of pre-training epochs are 400 and 1,600 in the former scenario or 300 and 800 in the latter scenario. We train all these models using 64 NVIDIA Tesla-V100 GPUs. For the base-level models, one pixel-supervised epoch and one CLIP-supervised epoch take about 2.7 and 4.7 minutes, respectively. For the large-level models, the numbers are 4.2 and 12.0 minutes, respectively. That said, a 1600-epoch pixel-supervised pre-training of iTPN-base/large takes around 75/115 hours. {\textcolor{black}{We use smooth-l1 loss function when using CLIP as supervision.}}
{\textcolor{black}{For Fast-iTPN, we follow the most configurations above. Differently, we pre-train all the Fast-iTPN series for 1600 epochs using CLIP-L as supervision and cosine similarity as loss function. }}

\begin{table}[t]
\centering
\setlength{\tabcolsep}{0.15cm}
\caption{Top-1 classification accuracy (\%) by fine-tuning the pre-trained models on ImageNet-1K. 
%Models of different levels and supervisions are compared.
}
\begin{tabular}{l|cc|cc|c}
\toprule
Method & \multirow{2}{*}{Arch.} & \multirow{2}{*}{Sup.} & \multirow{2}{*}{Eps.} & Param. & FT \\ 
\textit{at \textsf{base}-level} & {} & {} & {} & (M) & acc. \\

\midrule
BEiT~\cite{bao2021beit} & ViT-B & DALL-E & 400 & 86 & 83.2 \\
CAE~\cite{chen2022context} & ViT-B & DALL-E & 800 & 86 & 83.6 \\
PeCo~\cite{dong2021peco} & ViT-B & codebook & 300 & 86 & 84.5 \\
MaskFeat~\cite{maskfeat} & ViT-B & HOG & 800 & 86 & 84.0 \\
SimMIM~\cite{xie2022simmim} & ViT-B & pixel & 800 & 86 & 83.8 \\
SimMIM~\cite{xie2022simmim} & Swin-B & pixel & 800 & 88 & 84.0 \\
data2vec~\cite{data2vec} & ViT-B & pixel & 800 & 86 & 84.2 \\
ConvMAE~\cite{convmae}  & ConViT-B  & pixel  & 1600  & 88  & 85.0 \\
MAE~\cite{MAE2022} & ViT-B & pixel & 400 & 86 & 83.1 \\ 
MAE~\cite{MAE2022} & ViT-B & pixel & 1600 & 86 & 83.6 \\ 
HiViT~\cite{hivit} & HiViT-B & pixel & 800 & 66 & 84.2 \\ 
\rowcolor{Gray}
iTPN (ours) & HiViT-B & pixel & 400 & 79 & \textbf{85.1} \\
\rowcolor{Gray}
iTPN (ours) & HiViT-B & pixel & 1600 & 79 & \textbf{85.5} \\
\cmidrule(lr){1-1}\cmidrule(lr){2-2}\cmidrule(lr){3-3}\cmidrule(lr){4-4}\cmidrule(lr){5-5}\cmidrule(lr){6-6}
MVP~\cite{mvp} & ViT-B & CLIP-B & 300 & 86 & 84.4 \\
BEiT-v2~\cite{beitv2} & ViT-B  &CLIP-B  & 1600 & 86 & 85.5 \\
\rowcolor{Gray}
iTPN (ours) & HiViT-B & CLIP-B & 300 & 79 & \textbf{85.9} \\
\rowcolor{Gray}
iTPN (ours) & HiViT-B & CLIP-B & 800 & 79 & \textbf{86.2} \\
% Fast-iTPN (ours) & HiViT-B & CLIP-L & 1600 & 83 & \textbf{88.7} \\
\midrule\midrule
Method & \multirow{2}{*}{Arch.} & \multirow{2}{*}{Sup.} & \multirow{2}{*}{Eps.} & Param. & FT \\ 
\textit{at \textsf{large}-level} & {} & {} & {} & (M) & acc. \\
\midrule
BEiT~\cite{bao2021beit} & ViT-L & DALL-E & 800 & 307 & 85.2 \\
PeCo~\cite{dong2021peco} & ViT-L & codebook & 800 & 307 & 86.5 \\
MaskFeat~\cite{maskfeat} & ViT-L & HOG & 300 & 307 & 84.4 \\
MaskFeat~\cite{maskfeat} & ViT-L & HOG & 1600 & 307 & 85.7 \\
SimMIM~\cite{xie2022simmim} & Swin-L  & pixel  & 800  & 197  & 85.4 \\
SimMIM~\cite{xie2022simmim} & Swin-H & pixel & 800 & 658 & 85.7 \\
data2vec~\cite{data2vec} & ViT-L & pixel & 800 & 307 & 86.2 \\
CAE~\cite{chen2022context} & ViT-L & DALL-E & 1600 & 307 & 86.3 \\
MAE~\cite{MAE2022} & ViT-L & pixel & 1600 & 307 & 85.9 \\
HiViT~\cite{hivit} & HiViT-L & pixel & 1600 & 288 & 86.1 \\
\rowcolor{Gray}
iTPN (ours) & HiViT-L & pixel & 400 & 288 & \textbf{86.3} \\
\rowcolor{Gray}
iTPN (ours) & HiViT-L & pixel & 1600 & 288 & \textbf{86.7} \\
\cmidrule(lr){1-1}\cmidrule(lr){2-2}\cmidrule(lr){3-3}\cmidrule(lr){4-4}\cmidrule(lr){5-5}\cmidrule(lr){6-6}
MVP~\cite{mvp} & ViT-L/16 & CLIP-B & 300 & 307 & 86.3 \\
BEiT-v2~\cite{beitv2}  & ViT-L/16  & CLIP-B  & 300 & 307  & 86.6 \\
\rowcolor{Gray}
iTPN (ours) & HiViT-L/16 & CLIP-B & 300 & 288 & \textbf{87.0} \\
\rowcolor{Gray}
iTPN (ours) & HiViT-L/16 & CLIP-L & 300 & 288 & \textbf{87.8} \\
\rowcolor{Gray}
iTPN (ours) & HiViT-L/14 & CLIP-L & 300 & 288 & \textbf{88.0} \\
% iTPN (ours) & HiViT-L/16 & CLIP-L & 800 & 288 & \bf8x.x \\
\midrule
\multicolumn{6}{l}{ImageNet-22K + 384 input size} \\
\rowcolor{Gray}
iTPN (ours) & HiViT-L/16 & CLIP-L & 300 & 288 & \textbf{89.2} \\
\midrule
\multicolumn{6}{l}{ImageNet-22K + 512 input size} \\
\rowcolor{Gray}
\textcolor{black}{iTPN (ours)} & \textcolor{black}{HiViT-g/16} & \textcolor{black}{CLIP-L} & \textcolor{black}{300} & \textcolor{black}{1000} & \textcolor{black}{\textbf{89.5}} \\
% Fast-iTPN (ours) & HiViT-L/16 & CLIP-L & 1600 & 312 & \textbf{89.5} \\
\bottomrule
\end{tabular}
\label{tab:finetuning}
\end{table}

\begin{table}[t]
\centering
\vspace{-4pt}
\setlength{\tabcolsep}{0.015cm}
\caption{\textcolor{black}{Top-1 classification accuracy (\%) comparison with SOTA methods on ImageNet-1K. The \textsf{tiny/small/base}-level models report new records and the \textsf{large}-level model reaches state-of-the-art results.}}
\begin{tabular}{l|cc|cc|c|cc}
\toprule
\textcolor{black}{Method} & \multirow{2}{*}{\textcolor{black}{Dataset}} & \multirow{2}{*}{\textcolor{black}{Sup.}} & \multirow{2}{*}{\textcolor{black}{Eps.}} & \textcolor{black}{21K} & \textcolor{black}{Para.} & \textcolor{black}{Inp./Pat.} & \textcolor{black}{FT} \\ 
\textit{\textcolor{black}{at \textsf{tiny}-level}} & {} & {} & {} & \textcolor{black}{FT}  & \textcolor{black}{(M)} & \textcolor{black}{Size} & \textcolor{black}{acc.} \\
\midrule
\textcolor{black}{HiViT-T~\cite{hivit}}       & \textcolor{black}{IN1K}  & \textcolor{black}{pixel}   & \textcolor{black}{300}  & \textcolor{black}{N} &\textcolor{black}{19} & \textcolor{black}{224/16}  & \textcolor{black}{80.2}  \\
\textcolor{black}{CAEv2-T~\cite{caev2}}      & \textcolor{black}{IN1K}  & \textcolor{black}{CLIP-B}  & \textcolor{black}{300}  & \textcolor{black}{N}  &\textcolor{black}{22}  & \textcolor{black}{224/16}  & \textcolor{black}{82.8}  \\
\textcolor{black}{CAEv2-T $+$~\cite{caev2}}  & \textcolor{black}{IN1K}  & \textcolor{black}{CLIP-B}  & \textcolor{black}{300}  & \textcolor{black}{N}  &\textcolor{black}{22}  & \textcolor{black}{224/16}  & \textcolor{black}{83.1}  \\
\textcolor{black}{EVA02-T~\cite{fang2023eva02}} & \textcolor{black}{IN21K} & \textcolor{black}{EVA-CLIP-g}  & \textcolor{black}{240}  &\textcolor{black}{ N} &\textcolor{black}{22} & \textcolor{black}{336/14}  & \textcolor{black}{85.8}  \\
\textcolor{black}{InternImage-T~\cite{wang2023internimage}} & \textcolor{black}{IN1K}  & \textcolor{black}{label}  & \textcolor{black}{300}  & \textcolor{black}{N}  & \textcolor{black}{30}  & \textcolor{black}{224/-}  & \textcolor{black}{83.5}    \\
\rowcolor{Gray}
\textcolor{black}{iTPN-T}        & \textcolor{black}{IN1K}  & \textcolor{black}{pixel}   & \textcolor{black}{300}  & \textcolor{black}{N}  &\textcolor{black}{19}  & \textcolor{black}{224/16}  & \textcolor{black}{81.7}  \\
\rowcolor{Gray}
\textcolor{black}{iTPN-T}        & \textcolor{black}{IN1K}  & \textcolor{black}{CLIP-L}  &\textcolor{black}{ 300}  & \textcolor{black}{N}  &\textcolor{black}{19}  & \textcolor{black}{224/16}  & \textcolor{black}{84.2}  \\
\rowcolor{Gray}
\textcolor{black}{Fast-iTPN-T}   & \textcolor{black}{IN1K}  & \textcolor{black}{CLIP-L}  & \textcolor{black}{1600} & \textcolor{black}{N}  &\textcolor{black}{24}  & \textcolor{black}{224/16}  & \textcolor{black}{85.1}  \\
\rowcolor{Gray}
\textcolor{black}{Fast-iTPN-T}   & \textcolor{black}{IN1K}  & \textcolor{black}{CLIP-L}  & \textcolor{black}{1600} & \textcolor{black}{N}  &\textcolor{black}{24}  & \textcolor{black}{384/16}  & \textcolor{black}{86.2}   \\
\rowcolor{Gray}
\textcolor{black}{Fast-iTPN-T}   & \textcolor{black}{IN1K}  & \textcolor{black}{CLIP-L}  & \textcolor{black}{1600} & \textcolor{black}{N}  &\textcolor{black}{24}  & \textcolor{black}{512/16 } &\textcolor{black}{ 86.5}   \\
% Fast-iTPN (ours) & HiViT-B & CLIP-L & 1600 & 83 & \textbf{88.7} \\
\midrule\midrule
% Method & \multirow{2}{*}{Dataset} & \multirow{2}{*}{Sup.} & \multirow{2}{*}{Eps.} & 22K  & Inp./Pat. & FT \\ 
% \textit{at \textsf{small}-level} & {} & {} & {} & FT ? & Size & acc. \\
\multicolumn{7}{l}{\textit{\textcolor{black}{at \textsf{small}-level}}} \\
\midrule
\textcolor{black}{HiViT~\cite{hivit}}  & \textcolor{black}{IN1K}  &\textcolor{black}{ pixel}   & \textcolor{black}{300}  & \textcolor{black}{N} &\textcolor{black}{38} & \textcolor{black}{224/16}  & \textcolor{black}{83.5}  \\
% HiViT~\cite{hivit}  & IN1K  & CLIP-L  & 300  & N  & 224/16  & xx.x  \\
\textcolor{black}{InternImage-S~\cite{wang2023internimage}}  & \textcolor{black}{IN1K}  & \textcolor{black}{label} & \textcolor{black}{300}  & \textcolor{black}{N} & \textcolor{black}{50}  & \textcolor{black}{224/-}  &\textcolor{black}{84.2} \\
\rowcolor{Gray}
\textcolor{black}{iTPN}        & \textcolor{black}{IN1K}  & \textcolor{black}{pixel}   & \textcolor{black}{300}  & \textcolor{black}{N} &\textcolor{black}{38} & \textcolor{black}{224/16}  & \textcolor{black}{84.0}  \\
\rowcolor{Gray}
\textcolor{black}{iTPN }       & \textcolor{black}{IN1K}  & \textcolor{black}{CLIP-L}  & \textcolor{black}{300}  & \textcolor{black}{N} &\textcolor{black}{38} & \textcolor{black}{224/16}  & \textcolor{black}{85.4}  \\
\rowcolor{Gray}
\textcolor{black}{Fast-iTPN}   & \textcolor{black}{IN1K}  & \textcolor{black}{CLIP-L}  & \textcolor{black}{1600} & \textcolor{black}{N} &\textcolor{black}{40} & \textcolor{black}{224/16}  & \textcolor{black}{86.4}   \\
\rowcolor{Gray}
\textcolor{black}{Fast-iTPN}   & \textcolor{black}{IN1K}  & \textcolor{black}{CLIP-L}  & \textcolor{black}{1600} & \textcolor{black}{N} &\textcolor{black}{40} & \textcolor{black}{384/16}  & \textcolor{black}{86.95}   \\
\rowcolor{Gray}
\textcolor{black}{Fast-iTPN}   & \textcolor{black}{IN1K}  & \textcolor{black}{CLIP-L}  & \textcolor{black}{1600} & \textcolor{black}{N} &\textcolor{black}{40} & \textcolor{black}{512/16}  & \textcolor{black}{87.8}   \\

\midrule\midrule
% Method & \multirow{2}{*}{Dataset} & \multirow{2}{*}{Sup.} & \multirow{2}{*}{Eps.} & 22K  & Inp./Pat. & FT \\ 
% \textit{at \textsf{base}-level} & {} & {} & {} & FT ? & Size & acc. \\
\multicolumn{7}{l}{\textit{\textcolor{black}{at \textsf{base}-level}}} \\
\midrule

\textcolor{black}{HiViT~\cite{hivit}}  & \textcolor{black}{IN1K}  & \textcolor{black}{pixel}   & \textcolor{black}{300}  & \textcolor{black}{N}  &\textcolor{black}{66.4}  & \textcolor{black}{224/16}  & \textcolor{black}{84.6}  \\
% HiViT~\cite{hivit}  & IN1K  & CLIP-L  & 300  & N  & 224/16  & xx.x  \\
\textcolor{black}{CAE v2~\cite{caev2}} & \textcolor{black}{IN1K}  & \textcolor{black}{CLIP-B}  & \textcolor{black}{300}  & \textcolor{black}{N}  &\textcolor{black}{86}  & \textcolor{black}{224/16}  & \textcolor{black}{85.2}  \\
\textcolor{black}{CAE v2 $+$~\cite{caev2}}  & \textcolor{black}{IN1K}  & \textcolor{black}{CLIP-B}  & \textcolor{black}{300}  & \textcolor{black}{N}  &\textcolor{black}{86} & \textcolor{black}{224/16}  & \textcolor{black}{85.3}  \\
\textcolor{black}{EVA 02~\cite{fang2023eva02}} & \textcolor{black}{IN21K} & \textcolor{black}{EVA-CLIP-g } & \textcolor{black}{150}  & \textcolor{black}{N}  &\textcolor{black}{86}  & \textcolor{black}{196/14}  & \textcolor{black}{87.0}  \\
\textcolor{black}{EVA 02~\cite{fang2023eva02}} & \textcolor{black}{IN21K} & \textcolor{black}{EVA-CLIP-g}  & \textcolor{black}{150}  &\textcolor{black}{ N}  &\textcolor{black}{86}  & \textcolor{black}{448/14}  & \textcolor{black}{88.3}  \\
\textcolor{black}{EVA 02~\cite{fang2023eva02}} & \textcolor{black}{IN21K} & \textcolor{black}{EVA-CLIP-g}  & \textcolor{black}{150}  & \textcolor{black}{Y}  &\textcolor{black}{86}  & \textcolor{black}{448/14}  & \textcolor{black}{88.6}  \\
\textcolor{black}{InternImage-B~\cite{wang2023internimage}} & \textcolor{black}{IN1K}  & \textcolor{black}{label}  & \textcolor{black}{300}  & \textcolor{black}{N} & \textcolor{black}{97}  & \textcolor{black}{224/-}  & \textcolor{black}{84.9}   \\
\rowcolor{Gray}
\textcolor{black}{iTPN}        & \textcolor{black}{IN1K}  &\textcolor{black}{ pixel}   & \textcolor{black}{300}  & \textcolor{black}{N}  &\textcolor{black}{79}  & \textcolor{black}{224/16}  & \textcolor{black}{84.9}  \\
\rowcolor{Gray}
\textcolor{black}{iTPN}        & \textcolor{black}{IN1K}  & \textcolor{black}{CLIP-L}  & \textcolor{black}{300}  & \textcolor{black}{N}  &\textcolor{black}{79}  & \textcolor{black}{224/16}  & \textcolor{black}{87.0}  \\
\rowcolor{Gray}
\textcolor{black}{Fast-iTPN}   & \textcolor{black}{IN1K}  & \textcolor{black}{CLIP-L}  & \textcolor{black}{1600} & \textcolor{black}{N}  &\textcolor{black}{85}  & \textcolor{black}{224/16}  & \textcolor{black}{87.4}  \\
\rowcolor{Gray}
\textcolor{black}{Fast-iTPN}   & \textcolor{black}{IN1K}  & \textcolor{black}{CLIP-L}  & \textcolor{black}{1600} & \textcolor{black}{N}  &\textcolor{black}{85}  & \textcolor{black}{512/16}  &\textcolor{black}{ 88.5}   \\
\rowcolor{Gray}
\textcolor{black}{Fast-iTPN}   & \textcolor{black}{IN1K}  & \textcolor{black}{CLIP-L}  & \textcolor{black}{1600} &\textcolor{black}{ Y}  &\textcolor{black}{85}  & \textcolor{black}{512/16}  & \textcolor{black}{88.75}   \\

\midrule\midrule
% Method & \multirow{2}{*}{Arch.} & \multirow{2}{*}{Sup.} & \multirow{2}{*}{Eps.} & Param. & FT \\ 
% \textit{at \textsf{large}-level} & {} & {} & {} & (M) & acc. \\
\multicolumn{7}{l}{\textit{\textcolor{black}{at \textsf{large}-level}}} \\
\midrule
\textcolor{black}{BEiTv3~\cite{wang2022image}} &\textcolor{black}{M38M}  &\textcolor{black}{CLIP-L}  & \textcolor{black}{150}  & \textcolor{black}{N}  & \textcolor{black}{2000}  & \textcolor{black}{512/16}  & \textcolor{black}{89.6}   \\
\textcolor{black}{EVA 02~\cite{fang2023eva02}} & \textcolor{black}{IN21K} & \textcolor{black}{EVA-CLIP-g } & \textcolor{black}{150}  & \textcolor{black}{N} &\textcolor{black}{304} & \textcolor{black}{448/14}  & \textcolor{black}{89.6}  \\
\textcolor{black}{InternImage-XL~\cite{wang2023internimage}}  & \textcolor{black}{IN21K}  & \textcolor{black}{label} & \textcolor{black}{90}  &\textcolor{black}{Y} &\textcolor{black}{335} & \textcolor{black}{384/-}  & \textcolor{black}{88.0}\\
\textcolor{black}{InternImage-H~\cite{wang2023internimage}}   & \textcolor{black}{IN21K}  & \textcolor{black}{label} & \textcolor{black}{90}  &\textcolor{black}{Y} &\textcolor{black}{1080} & \textcolor{black}{640/-}  &\textcolor{black}{ 89.6}\\
\rowcolor{Gray}
\textcolor{black}{iTPN}     & \textcolor{black}{IN1K}    & \textcolor{black}{CLIP-L}   & \textcolor{black}{300}   & \textcolor{black}{N} &\textcolor{black}{288} & \textcolor{black}{256/16}  & \textcolor{black}{88.0} \\
\rowcolor{Gray}
\textcolor{black}{iTPN}     & \textcolor{black}{IN1K }   & \textcolor{black}{CLIP-L}   & \textcolor{black}{300}   & \textcolor{black}{Y} &\textcolor{black}{288} & \textcolor{black}{512/16}  & \textcolor{black}{89.2} \\
\rowcolor{Gray}
\textcolor{black}{Fast-iTPN}  & \textcolor{black}{IN1K}  & \textcolor{black}{CLIP-L}   & \textcolor{black}{1600}  & \textcolor{black}{N} &\textcolor{black}{312} & \textcolor{black}{640/16}  & \textcolor{black}{89.5}  \\
% iTPN (ours) & HiViT-L/16 & CLIP-L & 800 & 288 & \bf8x.x \\

% Fast-iTPN (ours) & HiViT-L/16 & CLIP-L & 1600 & 312 & \textbf{89.5} \\
\bottomrule
\end{tabular}
\label{tab:finetuning_fast_itpn}
\end{table}

\subsection{Image Classification}  
\label{exp:image classification}
% We report the results of the ImageNet-1K classification. Following the convention, we insert a linear classifier on top of \textcolor{black}{the last encoder block} and fine-tune the entire network. The number of epochs is $100$ for \textsf{base}-level models and $50$ for \textsf{large}-level models. We use the \textcolor{black}{AdamW} optimizer, with the initial learning rate being $5\times10^{-4}$ and $1\times10^{-3}$ for \textsf{base}-level and \textsf{large}-level models, respectively. The weight decay is $0.05$ and the batch size is $1\rm{,}024$. The number of warm-up epochs is $5$. The layer decay is set to be $0.55$ and $0.50$ for \textsf{base}-level and \textsf{large}-level models.
In this study, we present the results obtained from the ImageNet-1K classification task. Following established conventions, we incorporate a linear classifier on top of the final encoder block and conduct comprehensive fine-tuning of the entire network. \textbf{For iTPN}, the fine-tuning duration is set to 100 epochs for the base-level models and 50 epochs for the large-level models. To optimize the network, we employ the AdamW optimizer with initial learning rates of $5\times10^{-4}$ and $1\times10^{-3}$ for the base-level and large-level models respectively. A weight decay of 0.05 is applied, and a batch size of 1,024 is used during fine-tuning. Additionally, a warm-up period of 5 epochs is implemented, and layer decay is configured as 0.65 for the base-level models and 0.70 for the large-level models. 
{\textcolor{black}{\textbf{For Fast-iTPN}, we provide a total of 4 pre-trained models including Fast-iTPN-\textit{tiny} (Fast-iTPN-T) (24.2M), Fast-iTPN-\textit{small} (Fast-iTPN-S) (40.0M), Fast-iTPN-\textit{base} (Fast-iTPN-B) (85.2M), and Fast-iTPN-\textit{large} (Fast-iTPN-L) (312.3M). The optimization process for all models utilizes the AdamW optimizer with a weight decay value of 0.05 and a batch size of 1024. The initial learning rate is set to $1\times10^{-4}$ for the \textit{tiny/small/base} models and $5\times10^{-5}$ for Fast-iTPN-\textit{large}. Fine-tuning is performed for 100 epochs (with a warmup period of 5 epochs) for \textit{tiny/small} models, and for 20 epochs (with a warmup period of 3 epochs) for \textit{base/large} models. Regarding layer decay, the values are set to 0.80 for \textit{tiny}, 0.90 for \textit{small}, 0.90 for \textit{base}, and 0.95 for \textit{large} models.}}

Results of iTPN are summarized in Table~\ref{tab:finetuning}. One can see that iTPN achieves higher accuracy than existing methods on all tracks, \textit{i.e.}, using \textsf{base}-level or \textsf{large}-level backbones, with or without external supervision (\textit{i.e.}, CLIP~\cite{clip}). For example, using the \textsf{base}-level backbone, iTPN achieves an $85.1\%$ accuracy with only $400$ pre-training epochs, surpassing MAE~\cite{MAE2022} and HiViT~\cite{hivit} with $1\rm{,}600$ epochs. The accuracy of iTPN continues growing to $85.5\%$ with $1\rm{,}600$ pre-training epochs, which is on par with BEiT-v2~\cite{beitv2} that distilled knowledge from CLIP-B~\cite{clip} ($1\rm{,}600$ epochs), yet iTPN reports an $86.2\%$ accuracy with the supervision of CLIP ($800$ epochs). Similar situations occur when we use the \textsf{large}-level backbone, where the advantage of iTPN is a bit smaller due to the higher baseline. The best practice appears when an iTPN-L/$14$ model (\textit{i.e.}, patch size is adjusted to $14\times14$) is supervised by a CLIP-L teacher -- the classification accuracy, $88.0\%$, is the highest to date under fair comparisons. {\textcolor{black}{Using Imagenet-22K for intermediate fine-tuning further boosts the performance -- iTPN-L reaches 89.2 performance and the 1B model iTPN-g achieves 89.5 Top-1 accuracy.}}

{\textcolor{black}{Results of Fast-iTPN are summarized in Table~\ref{tab:finetuning_fast_itpn} (without using acceleration strategies). 
%We report the results of 4 Fast-iTPN series as shown.
Compared to iTPN, Fast-iTPN reaches higher performance. That is because Fast-iTPN gets more comprehensive pre-training and improved implementation details.
Compared to the state-of-the-art methods, the \textit{tiny/small/base} models of Fast-iTPN remain advantageous. Specifically, Fast-iTPN-T achieves 86.5\% top-1 performance with 24M parameters, which outperforms the previous best result (EVA02 uses more training data) by $0.7\%$  and also outperforms recent method InternImage~\cite{wang2023internimage}. 
Fast-iTPN-S achieves $87.8\%$ Top-1 accuracy, which is the best performance reported. Fast-iTPN-B model reaches $88.75\%$ accuracy, higher than the previous best model by $0.15\%$ with less computational cost and weaker supervision (compared to EVA02~\cite{fang2023eva02}). Fast-iTPN-L reaches a good performance of $89.5\%$, which is comparable to much larger models such as BEiTv3~\cite{wang2022image} and InternImage-XL/H~\cite{wang2023internimage}. 
%All the results show the potential of the iTPN/Fast-iTPN series and all the models will be released to the community. 
}}

\begin{table*}[t]
\centering
\setlength{\tabcolsep}{0.25cm}
\caption{Visual recognition results (\%) on COCO (object detection and instance segmentation, AP) and ADE20K (semantic segmentation, mIoU). Mask R-CNN (\textit{abbr.} MR, $1\times$/$3\times$) and Cascade Mask R-CNN (\textit{abbr.} CMR, $1\times$) are used on COCO, and UPerHead with $512\times512$ input is used on ADE20K. For the \textsf{base}-level models, each cell of COCO results contains object detection (box) and instance segmentation (mask) APs. For the \textsf{large}-level models, the accuracy of $1\times$ Mask R-CNN surpasses all existing methods. $^\dagger$: ConvMAE used a different setting from all other methods -- fine-tuning using ViTDet~\cite{VitDET2022} for $25$ epochs.}
\selectfont
\begin{tabular}{l|cc|cc|ccc|c}
\toprule
Method & \multirow{2}{*}{Arch.} & \multirow{2}{*}{Sup.} & \multirow{2}{*}{Eps.} & Param. & \multicolumn{3}{c|}{COCO} & ADE20K \\
\textit{at \textsf{base}-level} & {} & {} & {} & (M) & MR, $1\times$ & MR, $3\times$ & CMR, $3\times$ & UPerHead \\
\midrule
MoCo-v3~\cite{chen2021empirical} & ViT-B & pixel & 300 & 86 & 45.5/40.5 & -- & -- & 47.3 \\
BEiT~\cite{bao2021beit} & ViT-B & DALL-E & 400 & 86 & 42.1/37.8 & -- & -- & 47.1 \\
DINO~\cite{dino} & ViT-B & pixel & 400 & 86 & 46.8/41.5 & -- & -- & 47.2 \\
iBoT~\cite{zhou2021ibot} & ViT-B & pixel & 1600 & 86 & -- & -- & 51.2/44.2 & 50.0 \\
CAE~\cite{chen2022context} & ViT-B & DALL-E & 1600 & 86 & 50.0/44.0 & -- & -- & 50.2 \\
SimMIM~\cite{xie2022simmim} & Swin-B & pixel & 800 & 88 & -- & 52.3/-- & -- & 52.8  \\
MAE~\cite{MAE2022} & ViT-B & pixel & 1600 & 86 & 48.4/42.6 & -- & -- & 48.1 \\
ConvMAE~\cite{convmae} & ConViT-B & pixel & 1600 & 88 & -- & 53.2/47.1$^\dagger$ & -- & 51.7 \\
HiViT~\cite{MAE2022} & HiViT-B & pixel & 1600 & 66 & 49.5/43.8 & 51.2/44.2 & -- & 51.2 \\
MVP~\cite{mvp} & ViT-B & CLIP-B & 300 & 86 & -- & -- & 53.5/46.3 & 52.4 \\
% iTPN (ours) & HiViT-B & pixel & 1600 & 79 & \textbf{53.0/46.5} & \textbf{54.0/47.4} &  \textbf{56.0/48.5} & \textbf{53.5} \\
% iTPN (ours) & HiViT-B & CLIP-B & 800 & 79 & \textbf{53.3/46.7} & \textbf{54.2/47.5} &\textbf{56.1/48.6} & \textbf{54.7} \\
\rowcolor{Gray}
iTPN (ours) & HiViT-B & pixel & 1600 & 79 & \textbf{53.0}/\textbf{46.5} & \textbf{54.0}/\textbf{47.4} & \textbf{56.0}/\textbf{48.5} & \textbf{53.5} \\
\rowcolor{Gray}
iTPN (ours) & HiViT-B & CLIP-B & 800 & 79 & \textbf{53.2}/\textbf{46.6} & \textbf{54.1}/\textbf{47.5} & \textbf{56.1}/\textbf{48.6} & \textbf{54.7} \\
\midrule\midrule
Method & \multirow{2}{*}{Arch.} & \multirow{2}{*}{Sup.} & \multirow{2}{*}{Eps.} & Param. & \multicolumn{3}{c|}{COCO} & ADE20K \\
\textit{at \textsf{large}-level} & {} & {} & {} & (M) & object det. & instance seg. & \textit{schedule} & UPerHead \\
\midrule
% 		SimMIM~\cite{xie2021simmim}  &800  &Swin-L  & IN1K  & & &  \\
MAE~\cite{MAE2022} & ViT-L & pixel & 1600 & 307 & 54.0 & 47.1 & MR, $1\times$ & 53.6 \\
SimMIM~\cite{xie2022simmim} & Swin-L & pixel & 800 & 197 & 53.8 & -- & MR, $3\times$ & 53.5 \\
SimMIM~\cite{xie2022simmim} & SwinV2-H & pixel & 800 & 658 & 54.4 & -- & MR, $3\times$ & 54.2 \\
CAE~\cite{MAE2022} & ViT-L & pixel & 1600 & 304 & 54.5 & 47.6 & MR, $1\times$ & 54.7 \\
\rowcolor{Gray}
iTPN (ours) & HiViT-L & pixel & 1600 & 288 & \textbf{55.6} & \textbf{48.6} & MR, $1\times$ & \textbf{56.1} \\
\rowcolor{Gray}
iTPN (ours) & HiViT-L & CLIP-L & 300 & 288 & \textbf{55.2} & \textbf{48.2} & MR, $1\times$ & \textbf{57.7} \\
\midrule
\end{tabular}
\label{tab:downstream}
\end{table*}

\subsection{Detection and Segmentation}
\label{exp:detseg}
\textbf{COCO: object detection \& instance segmentation} We follow the configuration provided by CAE to evaluate the pre-trained models on the COCO dataset. We use Mask R-CNN implemented by MMDetection. We use the AdamW optimizer with a weight decay of 0.05. The standard 1× (12 epochs) and 3× schedules are applied, where the initial learning rate is $\mathbf{3 \times {10}^{-4}}$ and it decays by a factor of 10 after 3/4 and 11/12 of fine-tuning epochs. The layer-wise decay rate is set to be 0.90. We also try a 3× Cascade Mask R-CNN towards higher accuracy. We apply multi-scale training and single-scale testing.

Results are summarized in Table ~\ref{tab:downstream}. Compared to image classification, the advantages of iTPN become more significant because the pre-trained neck is reused, so that the fine-tuning stage only needs to initialize a task-specific head. For example, using a pixel-supervised base-level backbone, the 1× Mask R-CNN produces 53.0\% box AP, surpassing all other methods significantly (e.g., +4.6\% over MAE [25] and +3.0\% over CAE ). Compared to HiViT which did not pre-train the feature pyramid, iTPN claims a +1.7\% gain in box AP. The 1× Mask R-CNN results are even competitive among prior methods with a 3× Mask RCNN or a 3× Cascade Mask R-CNN. With these stronger heads, iTPN reports stronger numbers, e.g., the box/mask AP is 56.0\%/48.5\% using 3× Cascade Mask R-CNN, setting a new record with base-level models. The advantages persist when either CLIP supervision is introduced or the large-level backbone is used. Later, we will show that the benefits indeed come from pre-training the feature pyramid and loading it for downstream fine-tuning.

\textcolor{black}{The results of Fast-iTPN are summarized in Table~\ref{tab:downstream_fast_itpn} and we do not accelerate the inference (without using token migration and gathering) in this table to demonstrate the potential of Fast-iTPN. We use DINO~\cite{dino} detector for COCO object detection. Under 12 epoch training schedule, Fast-iTPN-T/S models achieve 54.1/55.8 mAP respectively, even surpassing Swin-B, ViTDet-B, and SwinV2-H models. At the base-level model, with training 12 epochs, Fast-iTPN reports 58.4 mAP, surpassing Swin-B/L, ViTDet-B, and iTPN-B/L models, matching EVA02-B model with pretraining on objects365. Fast-iTPN-B attains 61.8 mAP performance if pre-trained on the Objects365 dataset, which is the first model to breach $60+$ performance as far as we know. The Fast-iTPN-L model reports $58.8$ mAP, which also beats the counterparts. The dominant performance of these series demonstrates that Fast-iTPN has the potential to be a powerful backbone for downstream vision tasks.
}

\textbf{ADE20K: semantic segmentation}
We follow BEiT~\cite{bao2021beit} to build a UperHead~\cite{upernet} on top of the pre-trained backbone. We use the AdamW optimizer~\cite{adamw} and the learning rate is fixed as $3\times10^{-5}$. We fine-tune the model for a total of $160k$ iterations and the batch size is $16$. The input resolution is $512\times512$ and we do not use a multi-scale test. Results are summarized in Table~\ref{tab:downstream}. Again, iTPN reports the best accuracy in terms of mIoU. In particular, the pixel-supervised \textsf{base}/\textsf{large}-level models report $53.5\%$/$56.1\%$ mIoUs which surpass all the competitors substantially. Introducing CLIP supervision further improves both numbers by more than $1\%$, setting solid new records for both \textsf{base}-level and \textsf{large}-level models.

\textcolor{black}{Fast-iTPN results on ADE20K are summarized in Table~\ref{tab:downstream_fast_itpn}, where we do not use acceleration configurations like COCO. As shown, Fast-iTPN-T/S report 53.2/54.6 mIoU, which even surpasses Swin-B using UPerNet. Fast-iTPN-B attains 57.5 mIoU performance, beating the counterparts including EVA02. With pre-training on objects365, Fast-iTPN reports 58.2 mIoU, which is the best performance on base-level models as far as we know. The Fast-iTPN-L model achieves 58.7 mIoU, matching the Swin-L model that uses the ImageNet-21K datasets.}

% {\textcolor{black}{\textbf{Fast-iTPN results methods} %Above results provide the downstream comparison of iTPN models with other popular methods. In this part, we employ Fast-iTPN to demonstrate its potential in achieving high performance. 
% In Table~\ref{tab:downstream_fast_itpn}, Fast-iTPN is evaluated on object detection and semantic segmentation tasks. 
%
%Under same settings, Fast-iTPN outperforms state-of-the-art (SOTA) methods with significant margins. For example, 
% The Fast-iTPN-B model attains an outstanding mAP score of 61.8 on the COCO dataset, marking it as the initial base-level model to breach the 60+ performance threshold. 
%{\color{red}The Fast-iTPN-L model aso achieves remarkable results, with an mAP of 6x.x on COCO and an mIoU of 6x.x on ADE20K. These achievements underscore the considerable potential of Fast-iTPN.}}}

\begin{table*}[t]
\centering
\setlength{\tabcolsep}{0.02cm}
\caption{\textcolor{black}{Visual recognition comparison of different pre-trained models on COCO and ADE20K using state-of-the-art baseline towards better results.}}
\selectfont
\begin{tabular}{l|cc|cc|ccc|cc}
\toprule
\textcolor{black}{Method} & \textcolor{black}{Backbone Pre-training} & \multirow{2}{*}{\textcolor{black}{Sup.}} & \multirow{2}{*}{\textcolor{black}{Eps.}}  &  \textcolor{black}{Detection Pre-training} & \multicolumn{3}{c|}{\textcolor{black}{COCO}} & \multirow{2}{*}{\textcolor{black}{Segmentor}} &\multirow{2}{*}{\textcolor{black}{ADE20K, mIoU}} \\
\textit{\textcolor{black}{at \textsf{tiny/small}-level}}  & \textcolor{black}{Dataset} &{} &{}  & \textcolor{black}{Dataset}  & \textcolor{black}{Detector}  & \textcolor{black}{FT Eps.} & \textcolor{black}{mAP} & {}  \\
\midrule
\rowcolor{Gray}
\textcolor{black}{Fast-iTPN-T} &\textcolor{black}{IN1K}   & \textcolor{black}{CLIP-L}  & \textcolor{black}{1600}  & \textcolor{black}{N}  & \textcolor{black}{DINO}  & \textcolor{black}{12}   & \textcolor{black}{54.1}  & \textcolor{black}{MaskDINO}  & \textcolor{black}{53.2}  \\
\rowcolor{Gray}
\textcolor{black}{Fast-iTPN-S} &\textcolor{black}{IN1K}   & \textcolor{black}{CLIP-L}  & \textcolor{black}{1600 } & \textcolor{black}{N}  & \textcolor{black}{DINO}  & \textcolor{black}{12}   & \textcolor{black}{55.8}  & \textcolor{black}{MaskDINO}  & \textcolor{black}{54.6}  \\
\midrule\midrule
% Method & \multirow{2}{*}{Arch.} & \multirow{2}{*}{Sup.} & \multirow{2}{*}{Eps.} & Param. & \multicolumn{3}{c|}{COCO} & ADE20K \\
% \textit{at \textsf{large}-level} & {} & {} & {} & (M) & object det. & instance seg. & \textit{schedule} & UPerHead \\
\multicolumn{7}{l}{\textit{\textcolor{black}{at \textsf{base}-level}}} \\
\midrule
\textcolor{black}{Swin-B~\cite{Swin2021}}   & \textcolor{black}{IN21K} & \textcolor{black}{label} & \textcolor{black}{300} & \textcolor{black}{N}  & \textcolor{black}{HTC++}  & \textcolor{black}{36}  &  \textcolor{black}{56.4}  & \textcolor{black}{UPerNet}  & \textcolor{black}{51.6} \\
\textcolor{black}{Swin-B~\cite{Swin2021}}   & \textcolor{black}{IN21K} & \textcolor{black}{label} & \textcolor{black}{300} & \textcolor{black}{N}  & \textcolor{black}{DINO}   & \textcolor{black}{12}  &  \textcolor{black}{55.3}  & \textcolor{black}{UPerNet}  & \textcolor{black}{51.6} \\
\textcolor{black}{ViTDet-B~\cite{VitDET2022}}  & \textcolor{black}{IN1K}  & \textcolor{black}{pixel}  & \textcolor{black}{1600}  &\textcolor{black}{N}  &\textcolor{black}{ViTDet}  & \textcolor{black}{100}  & \textcolor{black}{54.0}  & - & - \\
\textcolor{black}{EVA02-B~\cite{fang2023eva02}}   & \textcolor{black}{IN21K}  & \textcolor{black}{EVA-CLIP-g}  & \textcolor{black}{150}  & \textcolor{black}{N}  & \textcolor{black}{Cas. MaskRCNN}  & \textcolor{black}{36} & \textcolor{black}{58.9} & \textcolor{black}{UPerNet} & \textcolor{black}{55.3} \\ 
\rowcolor{Gray}
\textcolor{black}{iTPN-B}      & \textcolor{black}{IN1K}   & \textcolor{black}{CLIP-L}  & \textcolor{black}{1600}  & \textcolor{black}{N}  & \textcolor{black}{DINO}  & \textcolor{black}{12}   & \textcolor{black}{56.1}  & \textcolor{black}{UPerNet}  & \textcolor{black}{55.3}  \\
\rowcolor{Gray}
\textcolor{black}{Fast-iTPN-B} &\textcolor{black}{IN1K}   & \textcolor{black}{CLIP-L}  & \textcolor{black}{1600}  & \textcolor{black}{N}  & \textcolor{black}{DINO } & \textcolor{black}{12}   & \textcolor{black}{58.4}  & \textcolor{black}{UPerNet}  & \textcolor{black}{55.8}  \\
\rowcolor{Gray}
\textcolor{black}{Fast-iTPN-B} &\textcolor{black}{IN1K}   & \textcolor{black}{CLIP-L}  & \textcolor{black}{1600}  & \textcolor{black}{N}  & \textcolor{black}{DINO}  & \textcolor{black}{12}   & \textcolor{black}{58.4}  & \textcolor{black}{MaskDINO}  & \textcolor{black}{57.5}  \\
\rowcolor{Gray}
\textcolor{black}{Fast-iTPN-B} &\textcolor{black}{IN1K}   & \textcolor{black}{CLIP-L}  & \textcolor{black}{1600}  & \textcolor{black}{O365 (v1)}  & \textcolor{black}{DINO}  & \textcolor{black}{6}   & \textcolor{black}{61.8}  & \textcolor{black}{MaskDINO}  & \textcolor{black}{58.2}  \\
\midrule\midrule
% Method & \multirow{2}{*}{Arch.} & \multirow{2}{*}{Sup.} & \multirow{2}{*}{Eps.} & Param. & \multicolumn{3}{c|}{COCO} & ADE20K \\
% \textit{at \textsf{large}-level} & {} & {} & {} & (M) & object det. & instance seg. & \textit{schedule} & UPerHead \\
\multicolumn{7}{l}{\textit{\textcolor{black}{at \textsf{large}-level}}} \\
\midrule
\textcolor{black}{Swin-L~\cite{Swin2021}}   & \textcolor{black}{IN21K} & \textcolor{black}{label} & \textcolor{black}{300} & \textcolor{black}{N}  & \textcolor{black}{HTC++}  & -  &  \textcolor{black}{57.1}  & \textcolor{black}{MaskDINO}  & \textcolor{black}{56.6} \\
% Swin-L~\cite{Swin2021}   & IN21K & label & 300 & O365  & DINO  & -  &  63.1  & MaskDINO  & - \\
\textcolor{black}{SwinV2-H~\cite{xie2022simmim}} & \textcolor{black}{IN1K} & \textcolor{black}{pixel} & \textcolor{black}{800} & \textcolor{black}{N}  & \textcolor{black}{Mask-RCNN}  & \textcolor{black}{36}  &  \textcolor{black}{54.4}  & \textcolor{black}{UPerNet}  & \textcolor{black}{54.2} \\
% EVA02-L~\cite{fang2023eva02}  & IN21K & EVA-CLIP-g  & 150  &N & Cascade Mask-RCNN$^\dagger$  & 36  & & UPerNet  & \\
\textcolor{black}{ViTDet-L~\cite{VitDET2022}}  & \textcolor{black}{IN1K}  &\textcolor{black}{ pixel}  & \textcolor{black}{1600}  &\textcolor{black}{ N}  &\textcolor{black}{ViTDet}  & \textcolor{black}{100}  & \textcolor{black}{57.6}  & - & - \\
% ViTDet-H~\cite{VitDET2022}  & IN1K  & pixel  & 1600  & N  &ViTDet  & 100  & 58.7  & - & - \\
% MViTv2-L~\cite{mvit_v2}  & IN1K   & label   & 300   & N  & Cascade Mask-RCNN  & 36  & 56.9  & - & - \\
% MViTv2-H~\cite{mvit_v2}  & IN1K   & label   & 300   & N  & Cascade Mask-RCNN  & 36  & 57.1  & - & - \\
\rowcolor{Gray}
\textcolor{black}{iTPN-L}     & \textcolor{black}{IN1K}   & \textcolor{black}{pixel}  & \textcolor{black}{1600}  & \textcolor{black}{N}  & \textcolor{black}{DINO}  & \textcolor{black}{12}   & \textcolor{black}{57.7}  & \textcolor{black}{MaskDINO}  & \textcolor{black}{57.9} \\
\rowcolor{Gray}
\textcolor{black}{Fast-iTPN-L} &\textcolor{black}{IN1K}   & \textcolor{black}{CLIP-L}  & \textcolor{black}{1600}  & \textcolor{black}{N}  & \textcolor{black}{DINO}  & \textcolor{black}{12}   & \textcolor{black}{58.8}  & \textcolor{black}{MaskDINO}  & \textcolor{black}{58.7}  \\
% Fast-iTPN-L &IN1K   & CLIP-L  & 1600  & O365  & DINO  & 12   & xx.x  & MaskDINO  & xx.x  \\
% Fast-iTPN-L &IN1K   & CLIP-L  & 1600  & O365  & DINO  & 36   & xx.x  & MaskDINO  & xx.x  \\
\midrule
\end{tabular}
\label{tab:downstream_fast_itpn}
\end{table*}

% \subsection{\textcolor{black}{Performance-Speed Tradeoff}}

% \textcolor{black}{
% %We present the empirical findings along with a comprehensive trade-off analysis $w.r.t.$ the interplay between performance metrics and inference speed within the Fast-iTPN framework. 
% This experiment is performed on the MS COCO dataset~\cite{COCO2014}, using the ViTDet detector~\cite{VitDET2022}. Training epoch is adjusted to 36 epochs and the input sizes are set to $1024\times1024$.
% %
% The results are summarized in Figure~\cite{fig:perf_speed_trade}. Without token migration and gathering tokens, Fast-iTPN demonstrates an mAP of $xx.x\%$ alongside the throughput of xx.x images/s.
% %
% With $10\%$, $20\%$, $30\%$, and $40\%$ token migration ratios, Fast-iTPN respectively achieves $xx.x\%$, $xx.x\%$, $xx.x\%$, and $xx.x\%$ and inference speeds xx.x, xx.x, xx.x, and xx.x images/s. 
% %
% %This is achieved through the selective dropping of $10\%$, $20\%$, $30\%$, and $40\%$ of tokens.
% With window-leader, Fast-iTPN achieves an mAP of $xx.x\%$ alongside the inference speed of xx.x images/s. This corresponds to a noteworthy improvement in inference speed by approximately 50\%.
% %
% With token migration and window-leader, Fast-iTPN exhibits an enhanced inference speed, xx.x images/s, and $0.2\%$ performance drop. 
% %
% It is worth noting that it remains outperforming ViT by $2.3\%$ while operating at a 20\% higher speed.
% }

\begin{table}
\caption{\textcolor{black}{Evaluation of gathering tokens for rotated object detection on DOTA~\cite{xia2018dota}. The models are fine-tuned for $1\times$ using Faster R-CNN. GT: gathering tokens, WMSA: window self-attention, RPE: relative position encoding for WMSA and Cross Attention. HRPE: relative position encoding only for WMSA.}}
\label{tab:global_token}
\setlength{\tabcolsep}{0.08cm}
\centering
\begin{tabular}{l|cccc|cc}
\toprule
&\textcolor{black}{GT} & \textcolor{black}{WMSA} & \textcolor{black}{RPE} & \textcolor{black}{HRPE} & \textcolor{black}{mAP (\%)} & \textcolor{black}{speed (imgs/s)} \\
\midrule
\textcolor{black}{W/o Speedup}  &\textcolor{black}{\xmark} & \textcolor{black}{\xmark} & \textcolor{black}{\xmark} & \textcolor{black}{\xmark} & \textcolor{black}{79.3} & \textcolor{black}{17.3} \\
&\textcolor{black}{\xmark} & \textcolor{black}{\checkmark} & \textcolor{black}{\xmark} & \textcolor{black}{\xmark} & \textcolor{black}{77.7} & \textcolor{black}{28.4} \\
&\textcolor{black}{\xmark} & \textcolor{black}{\checkmark} & \textcolor{black}{\xmark} & \textcolor{black}{\checkmark} & \textcolor{black}{77.9} & \textcolor{black}{27.8} \\
&\textcolor{black}{\checkmark} & \textcolor{black}{\checkmark} & \textcolor{black}{\xmark} & \textcolor{black}{\xmark} & \textcolor{black}{78.6} &\textcolor{black}{ 28.7} \\
&\textcolor{black}{\checkmark} & \textcolor{black}{\checkmark} & \textcolor{black}{\checkmark} & \textcolor{black}{\xmark} &\textcolor{black}{ 77.8} & \textcolor{black}{25.9} \\
\textcolor{black}{W/ Speedup} &\textcolor{black}{\checkmark} & \textcolor{black}{\checkmark} & \textcolor{black}{\xmark} & \textcolor{black}{\checkmark} & \textcolor{black}{79.0} &\textcolor{black}{ 26.0} \\
\bottomrule
\end{tabular}
\end{table}

\subsection{Ablation Study}

\subsubsection{\textcolor{black}{Token Migration Ratio}}
% \textcolor{black}{
% In Figure ~\ref{tab:perf_speed_trade}, we evaluate the impact of ratios (\textbf{R}) of migrated tokens. When \textbf{R} = $10\%$, the inference speed of the model increased by $5\%$ without performance drop. When setting\textbf{R} = $4\%$, the inference speed increases by $20\%$ with $0.3\%$ mAP drop. In the following experiments, we set\textbf{R} = $40\%$ to balance speed and performance.}
\textcolor{black}{
%We present the empirical findings along with a comprehensive trade-off analysis $w.r.t.$ the interplay between performance metrics and inference speed within the Fast-iTPN framework. 
This experiment is performed on the MSCOCO dataset~\cite{COCO2014}, using the MaskRCNN detector~\cite{MaskRCNN2017}. Training epoch is 12 epochs using Fast-iTPN-B model.
The results are summarized in Figure~\ref{fig:perf_speed_trade}. Without token migration and gathering tokens (baseline), Fast-iTPN demonstrates an mAP of $53.4\%$. Replacing MLP with ConvMLP led to a $10\%$ speed improvement without performance loss.
With $10\%$, $20\%$, $30\%$, and $40\%$ token migration ratios, Fast-iTPN respectively achieves $53.4\%$, $53.3\%$, $53.2\%$, and $53.0\%$ mAPs and accelerate inference speed to $15\%$, $20\%$, $23\%$, and $27\%$. 
%
%This is achieved through the selective dropping of $10\%$, $20\%$, $30\%$, and $40\%$ of tokens.
Applying window attention causes a $0.6\%$ performance drop but $90\%$ speed improvement. 
With gathering tokens, Fast-iTPN achieves an mAP of $52.7\%$ alongside the inference improvement of approximately $85\%$.
With token migration and gathering tokens (further using relative position embedding), Fast-iTPN totally exhibits an enhanced inference speed by $70\%$, and $0.5\%$ performance drop. 
It is worth noting that it remains outperforming ViT by $2.3\%$ while operating at a 20\% higher speed.
}

\subsubsection{Gathering Tokens} 
\textcolor{black}{
In Figure~\ref{fig:perf_speed_trade}, we have shown the effect of gathering tokens.
In Table~\ref{tab:global_token}, we further evaluate the impact of employing gathering tokens in rotated object detection using the DOTA dataset~\cite{xia2018dota}. Fast-iTPN-B achieves a $79.3\%$ mAP after fine-tuning for $12$ epochs using the Faster R-CNN detector~\cite{FasterRCNN2015}. Utilizing the window attention technique results in a $1.6\%$ decrease in mAP performance but provides a 64\% increase in speed. Gathering tokens contributes to a $0.9\%$ gain in mAP without reducing inference speed. The introduction of relative position encoding for window attention results in a $79.0\%$ mAP performance, which incurs only a marginal $0.25\%$ ($\sim 0.3\%$) mAP performance decline while delivering a $50\%$ speedup.
}

\subsubsection{ConvMLP \& MFM}
We investigate the technical details of integral pre-training, in particular, using ConvMLP in the feature pyramid and applying MFM for multi-stage supervision. As shown in Table 8, both ConvMLP and MFM contribute individually to recognition accuracy, meanwhile, integrating them yields even better recognition performance.

\subsection{Visualization Analysis}
\label{exp:sec:analysis}
\begin{figure*}[!t]
\centering
\includegraphics[width=18cm]{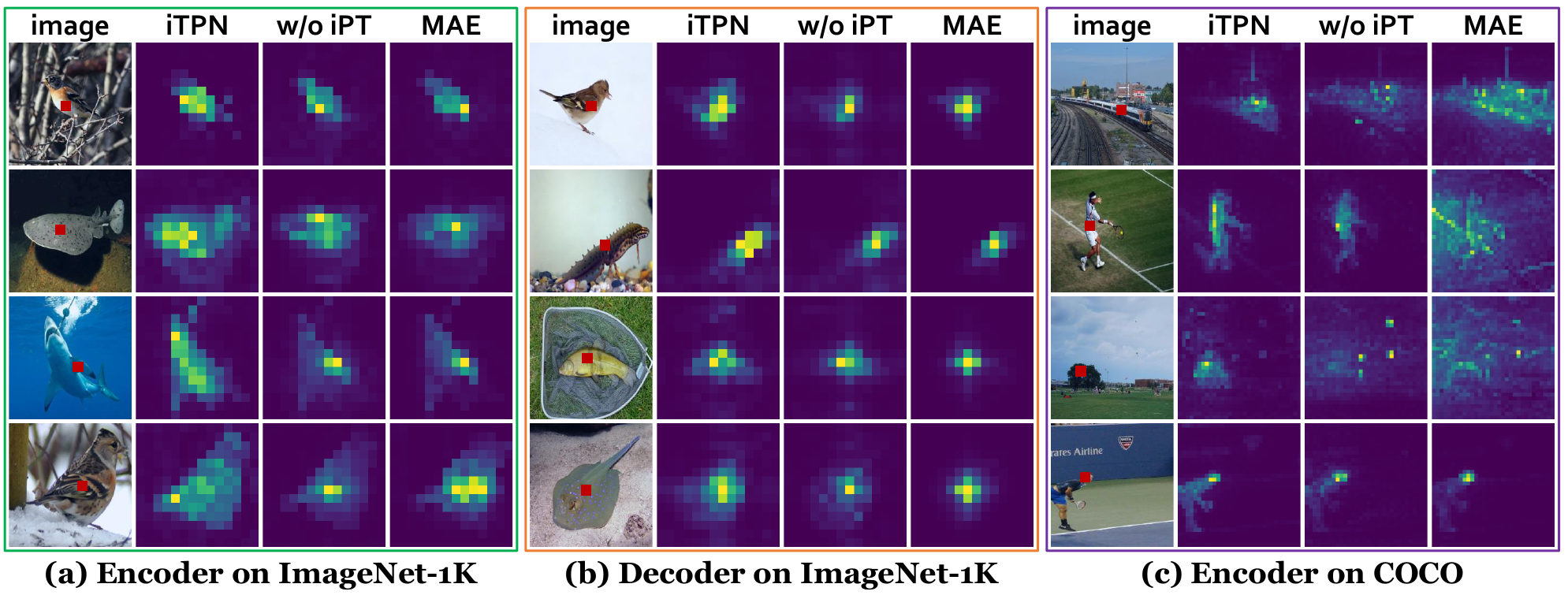}
\caption{A comparison between the attention maps generated by iTPN, the variant without integral pre-training (w/o iPT), and the MIM baseline (MAE~\cite{MAE2022}). In each case, the red block indicates the query token, and the attention map between the query and other tokens at the corresponding transformer block is shown. We use $224\times224$ input images in (a), (b), and $512\times512$ images in (c).}
\label{fig:attentions}
\end{figure*}

In Figure ~\ref{fig:attentions}, we visualize the attention maps generated by iTPN and baseline methods. (1) On the encoder, iTPN shows the advantage of detecting complete objects on ImageNet and concentrating on the chosen object on COCO. Such ability arises because iTPN forces the model to preserve richer visual features (multi-scale feature maps), which facilitates better recognition results in the downstream tasks. (2) On the decoder, iTPN can still realize the semantic relationship between tokens, resulting in better reconstruction results (Figure 4). We owe such benefits to the pre-trained neck that aggregates multi-stage visual features.

 \textbf{Ablative studies}\quad
Throughout this part, we use the $400$-epoch pixel-supervised model for diagnosis. We first ablate the benefit of integral pre-training. As shown in Table~\ref{tab:ablation1}, jointly optimizing the backbone and neck leads to higher recognition accuracy on all datasets including ImageNet-1K, COCO, and ADE20K. Beyond this point, loading the pre-trained feature pyramid (neck) further improves the recognition accuracy on COCO and ADE20K. This validates that the backbone itself is strengthened by iTPN, and thus it can be transferred to downstream tasks independently of the neck.

\begin{table}
\caption{Ablations on whether the model is integrally pre-trained (iPT) and whether the feature pyramid is loaded for detection and segmentation. Fine-tuning on ImageNet-1K does not involve loading the pyramid. The numbers are in \% for classification accuracy, box AP, and mIoU. The models are pre-trained for $400$ epochs. For COCO, $1\times$ Mask R-CNN is used and box AP is reported.}
\label{tab:ablation1}
\centering
\begin{tabular}{cc|ccc}
\toprule
iPT & loaded & ImageNet-1K & COCO & ADE20K \\
\midrule
\xmark & -- & 84.4 & 50.6 &51.5   \\
\checkmark & \xmark & \multirow{2}{*}{\textbf{85.1}} & 51.5 & 51.8 \\
\checkmark & \checkmark & {} & \textbf{52.1} & \textbf{52.2} \\
\bottomrule
\end{tabular}
\end{table}

\begin{table}
\caption{Ablations on C-MLP and MFM. The settings remain the same as in Table~\ref{tab:ablation1}. $^\ast$ indicates that convolution is used (instead of C-MLP) in the backbone and feature pyramid, which leads to lower performance.}
\label{tab:ablation2}
\centering
\begin{tabular}{cc|ccc}
\toprule
C-MLP & MFM & ImageNet-1K & COCO & ADE20K \\
\midrule
\xmark$^\ast$ & \xmark & 84.3 & 49.8 & 50.0 \\
\xmark$^\ast$ & \checkmark & 84.6 & 50.8 & 50.7 \\
\checkmark & \xmark & 84.9 & 51.8 & 51.8 \\
\checkmark & \checkmark & \textbf{85.1} & \textbf{52.1} & \textbf{52.2} \\
\bottomrule
\end{tabular}
\vspace{-0.0cm}
\end{table}

Next, we investigate the technical details of integral pre-training, in particular, using the channel-wise multi-layer perceptron (C-MLP) in the feature pyramid and applying masked feature modeling (MFM) for multi-stage supervision. As shown in Table~\ref{tab:ablation2}, both C-MLP and MFM contribute individually to recognition accuracy, meanwhile, integrating them yields even better recognition performance.

% 2. visualization -- important, (1) Fig 3 that shows the comparison between attentions generated by iTPN and other methods; (2) Fig xx that shows the comparison between different stages -- I guess this is important for downstream tasks, especially for instance-related tasks -- the ideal case is to show visualization on both ImageNet and COCO images for the benefit (complementariness) of multi-stage attentions: for ImageNet, combining multi-stage attention for whole object identification; for COCO, combining multi-stage attention for better detection of objects in different scales

\begin{figure}[!t]
\centering
\includegraphics[width=.99\linewidth]{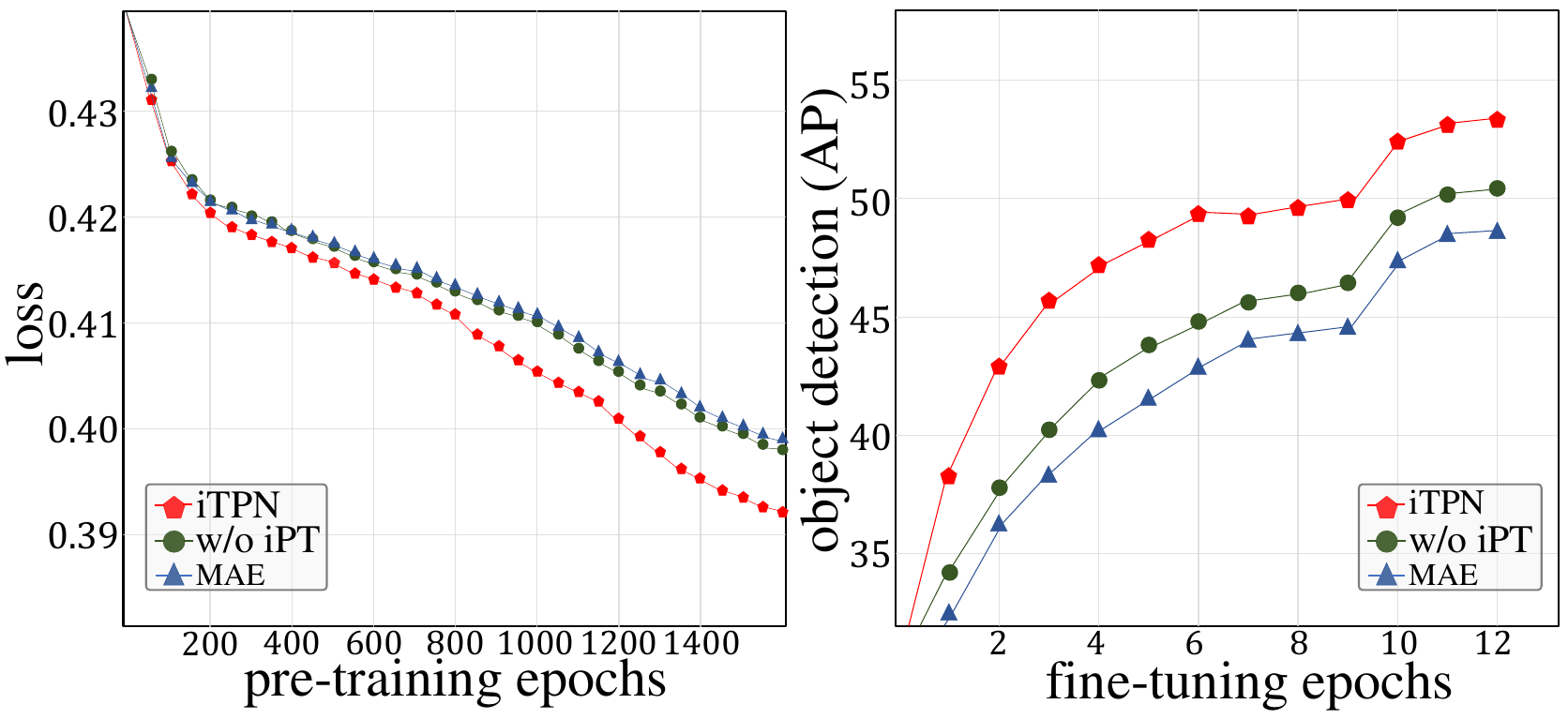}
\caption{\textbf{Left}: the comparison of reconstruction loss values of different frameworks. \textbf{Right}: the comparison of convergence speed in terms of box AP on COCO when the pre-trained models are fine-tuned with Mask R-CNN for $12$ epochs ($1\times$).}
\label{fig:analysis}
\end{figure}

 \textbf{Visualization}\quad
In Figure~\ref{fig:attentions}, we visualize the attention maps generated by iTPN and baseline methods. \textbf{(1)} On the encoder, iTPN shows the advantage of detecting complete objects on ImageNet and concentrating on the chosen object on COCO. Such ability arises because iTPN forces the model to preserve richer visual features (multi-scale feature maps), which facilitates better recognition results in the downstream tasks. (2) On the decoder, iTPN can still realize the semantic relationship between tokens, resulting in better reconstruction results (Figure~\ref{fig:analysis}). We owe such benefits to the pre-trained neck that aggregates multi-stage visual features.

The benefits brought by more complete attention can be quantified using two-fold experiments shown in Figure~\ref{fig:analysis}. \textbf{(1)} In the left part, we observe that iTPN achieves better reconstruction results (\textit{i.e.}, lower reconstruction loss values). Note that simply using a hierarchical vision transformer (with multi-scale feature maps) does not improve reconstruction, implying that integral pre-training is the major contributor. \textbf{(2)} In the right part, we show that better depiction of objects helps downstream visual recognition tasks (\textit{e.g.}, object detection) to converge faster and achieve a higher upper-bound -- this aligns with the outstanding accuracy on COCO (see Section~\ref{exp:detseg}). Integrating these analyses, we conclude that iTPN successfully transfers the benefits from upstream pre-training (reconstruction) to downstream fine-tuning (recognition), completing the entire chain.

\section{Conclusion}

We proposed an integrated framework for pre-training hierarchical vision transformers. The core contribution lies in a unified formulation that uses a feature pyramid for both reconstruction and recognition so that the transfer gap between pre-training and fine-tuning is maximally reduced. Besides, a masked feature modeling (MFM) task is designed to complement masked image modeling (MIM) to better optimize the feature pyramid. The pre-trained iTPNs report superior recognition in a few popular visual recognition tasks. 
\textcolor{black}{We further updated iTPN to Fast-iTPN, an enhanced and more efficient representation model tailored for downstream tasks.}
Our work clearly enlightens a future direction -- designing a unified framework for upstream and downstream visual representation learning.

\section*{Acknowledgments}
This work was supported by National Natural Science Foundation of China (NSFC) under Grant 62225208 and 62171431. 

\ifCLASSOPTIONcaptionsoff
  \newpage
\fi

\bibliographystyle{ieeetr}
\bibliography{IEEEabrv,egbib}

\end{document}